\documentclass[letterpaper]{article} % DO NOT CHANGE THIS
\usepackage{aaai2026}  % DO NOT CHANGE THIS
\usepackage{times}  % DO NOT CHANGE THIS
\usepackage{helvet}  % DO NOT CHANGE THIS
\usepackage{courier}  % DO NOT CHANGE THIS
\usepackage[hyphens]{url}  % DO NOT CHANGE THIS
\usepackage{graphicx} % DO NOT CHANGE THIS
\usepackage{natbib}  % DO NOT CHANGE THIS AND DO NOT ADD ANY OPTIONS TO IT
\usepackage{caption} % DO NOT CHANGE THIS AND DO NOT ADD ANY OPTIONS TO IT
\usepackage{algorithm}
\usepackage{algpseudocode}

\usepackage{newfloat}
\usepackage{listings}
\DeclareCaptionStyle{ruled}{labelfont=normalfont,labelsep=colon,strut=off} % DO 
\floatstyle{ruled}
\newfloat{listing}{tb}{lst}{}
\floatname{listing}{Listing}

\usepackage{romannum}
\usepackage{amsmath}
\usepackage{amssymb}
\usepackage{mathtools}
\usepackage{amsthm}
\usepackage{tikz} % nice language for creating drawings and diagrams
\usepackage{amsfonts,bm}
\usepackage{booktabs}
\usepackage{multirow} 
\usepackage{enumitem}
\usepackage{graphicx}
\usepackage{subfigure}
\usepackage{cite}
\usepackage[capitalize,noabbrev]{cleveref}
\usepackage{xcolor}

\xdefinecolor{blueZ}{RGB}{0,77,163}
\xdefinecolor{redZ}{RGB}{232,61,117}

\newtheorem{theorem}{Theorem}

\newtheorem{lemma}{Lemma}

\theoremstyle{definition}

\newtheorem{assumption}{Assumption}
\newtheorem{remark}{Remark}

\usepackage{pifont}
\newcommand{\cmarkgreen}{\textcolor{green!70!black}{\ding{51}}} % green check
\newcommand{\xmarkred}{\textcolor{red}{\ding{55}}}            % red cross

\title{Streaming Generated Gaussian Process Experts for Online Learning and Control: Extended Version}
\author {
Zewen Yang\textsuperscript{\rm 1}, 
Dongfa Zhang\textsuperscript{\rm 1}, 
Xiaobing Dai\textsuperscript{\rm 2}, 
Fengyi Yu\textsuperscript{\rm 1}, 
Chi Zhang\textsuperscript{\rm 3}, 
Bingkun Huang\textsuperscript{\rm 1}, 
Hamid Sadeghian\textsuperscript{\rm 1}, 
Sami Haddadin\textsuperscript{\rm 4}
\nocopyright
}
\affiliations {
\textsuperscript{\rm 1}Chair of Robotics and Systems Intelligence, Munich Institute of Robotics and Machine Intelligence, Technical University of Munich\\
\textsuperscript{\rm 2} Chair of Information-oriented Control, School of Computation, Information and Technology, Technical University of Munich\\
\textsuperscript{\rm 3}Department of Informatics, School of Computation, Information and Technology, Technical University of Munich\\
\textsuperscript{\rm 4}Mohamed bin Zayed University of Artificial Intelligence\\
zewen.yang@tum.de
}

\begin{document}

\maketitle

\begin{abstract}
	Gaussian Processes (GPs), as a nonparametric learning method, offer flexible modeling capabilities and calibrated uncertainty quantification for function approximations. 
	Additionally, GPs support online learning by efficiently incorporating new data with polynomial-time computation, making them well-suited for safety-critical dynamical systems that require rapid adaptation. 
	However, the inference and online updates of exact GPs, when processing streaming data, incur cubic computation time and quadratic storage memory complexity, limiting their scalability to large datasets in real-time settings. 
	In this paper, we propose a \underline{s}treaming \underline{k}ernel-induced progressivel\underline{y} generated expert framework of \underline{G}aussian \underline{p}rocesses (SkyGP) that addresses both computational and memory constraints by maintaining a bounded set of experts, while inheriting the learning performance guarantees from exact Gaussian processes. 
	Furthermore, two SkyGP variants are introduced, each tailored to a specific objective, either maximizing prediction accuracy (SkyGP-Dense) or improving computational efficiency (SkyGP-Fast).
	The effectiveness of SkyGP is validated through extensive benchmarks and real-time control experiments demonstrating its superior performance compared to state-of-the-art approaches.
\end{abstract}

\begin{links}
    \link{Code}{https://github.com/Zewen-Yang/SkyGP}
\end{links}

\section{Introduction}
\label{sec_intro}
Real-time learning has become essential for modeling the dynamics of physical systems, where machine learning models must be continuously updated to adapt to changing environments while satisfying the requirements of responsiveness and safety. 
This capability is especially critical for autonomous systems, such as underwater vehicles, aerial drones, and healthcare robots, that operate in complex and safety-critical environments~\cite{yang_EAAI2025_safe}.
In such scenarios, the ability to learn accurate models online and seamlessly integrate them into control loops is key to ensuring robust and efficient operation~\cite{dai_L4DC2023_can}.

Gaussian Processes (GPs) provide a powerful nonparametric framework for modeling dynamical systems, particularly in safety-critical applications, due to their ability to quantify uncertainty~\cite{Wachi_2018AAAI_safe}. 
However, despite their modeling flexibility, standard GP inference suffers from severe scalability limitations.
It scales cubically with the number of data points, requiring $\mathcal{O}(N^3)$ time and $\mathcal{O}(N^2)$ memory for $N$ training data. 
This scalability bottleneck makes conventional GPs impractical for online applications involving continuously streaming data or long-term deployments.

To mitigate these limitations, a variety of scalable approximation techniques have been developed. 
Among global approximations, sparse Gaussian Processes introduce a set of $M \ll N$ inducing points to efficiently summarize the training data, reducing the complexity to $\mathcal{O}(NM^2)$ for training and $\mathcal{O}(M^2)$ for prediction. Notable methods include fully independent training conditional approximation~\cite{Snelson_NeurIPS2005_Sparse}, variational free energy~\cite{Naish_NeurIPS2007_Generalized}, and latent projection techniques~\cite{Reeb_NeurIPS2018_Learning}. 
While streaming variants such as incremental sparse spectrum GP (ISSGP)~\cite{Arjan_NN2013_Real} and streaming sparse GP (SSGP)~\cite{bui_NeurIPS2017_Streaming} allow for incremental online updates, these frequently require computationally expensive optimization at each update, which can undermine their practical use in latency-sensitive settings. 
Moreover, these methods typically forgo guarantees on prediction error bounds, sacrificing reliability in critical applications.

Distributed Gaussian Processes (DGPs) provide complementary scalability by partitioning data across multiple processors or agents for parallel processing. Within this paradigm, mixture-of-experts (MoE) approaches~\cite{Tresp_NeurIPS2000_Mixtures, Yuan_NeurIPS2008_Variational,Trapp_AISTATS2020_Deep} aggregate predictions from independently trained GP experts using fixed or learned weights. 
Extensions such as product-of-experts (PoE)~\cite{Cohen_ICML2020_Healing}, generalized PoE (gPoE)~\cite{cao2015generalizedproductexpertsautomatic}, and correlated PoE with sparse GPs~\cite{Scheurch_ML2023_Correlated} have been proposed to better utilize uncertainty in aggregation. 
The Bayesian committee machine (BCM)~\cite{tresp_NC2000_bayesian} and its robust variant (rBCM)~\cite{Deisenroth_ICML2015_Distributed,liu_ICML2018_Generalized} explicitly integrate the GP prior to mitigate overconfidence.  
To address real-time learning problems within distributed frameworks, \citet{lederer_ICML2021_Gaussian} 
proposed LoG-GP, which incrementally constructs a tree-structured ensemble of GP experts offering a scalable solution for DGPs.
However, the reliance on partitioning along a single dimension limits effectiveness in high-dimensional spaces. 
The structure of DGPs enables parallel and distributed deployment, with each GP model hosted on a separate computational node or agent. 
This is widely employed in multi-agent systems, where agents can independently operate local GPs and collaborate through a communication network for cooperative learning~\cite{yang_CDC2021_Distributed,yang_AAMAS2024_whom,yang_ACC2024_cooperative,Lederer_TAC2024_cooperative}.
Nevertheless, these methods overlook online learning requirements, particularly the need to efficiently incorporate new observations and adapt to nonstationary environments~\cite{ Yuan_TAC2024_Lightweight,dai_EJC2024_Decentralized,dai_TNNLS2025_Cooperative}. 
While~\citet{yang_AAAI2025_Asynchronous} tackle asynchronous communication, their computation-aware random splitting strategy fails to exploit spatial or temporal correlations in the streaming data.

These limitations motivate our streaming kernel-induced progressively generated expert framework for Gaussian processes (SkyGP), which handles non-stationary, streaming data by dynamically allocating GP experts based on kernel similarity and temporal recency.
Our main contributions:
\begin{itemize}
	\item We propose a progressive expert generation strategy that leverages kernel-induced centers to determine whether incoming data should be incorporated into an existing expert model or used to initialize a new one. 
	SkyGP enables dynamic partitioning of streaming data and addresses the limitations of the state-of-the-art (SOTA) approaches.
	
	\item We develop a time-aware and configurable expert aggregation framework that incorporates temporal weighting to effectively manage the generated GP experts. 
	The framework adapts to system constraints, such as memory and computational budgets, while ensuring bounded complexity during both training and inference. 
	
	\item We provide a learning-based policy for dynamical system control tasks with a rigorous theoretical analysis. 
	The main theorem reveals the relationship between model uncertainty and control performance.
	
	\item We validate the proposed approach on real-world benchmark datasets and real-time control tasks. Extensive experiments demonstrate that the Sky-GP outperforms SOTA methods in terms of prediction accuracy, computational efficiency, and closed-loop control performance.
\end{itemize}

%%%%%%%%%%%%%%%%%%%%%%%%%%%%%%%%%%%%%%%%%%%%%%%%%%%%%%%%%%%%%%%%
%%%%%%%%%%%%%%%%%%%%%%%%%%%%%%%%%%%%%%%%%%%%%%%%%%%%%%%%%%%%%%%%
\section{Problem Statement and Preliminaries}
\label{sec_problem}
This paper investigates the problem of online function approximation in a streaming data environment for learning and control tasks.
The objective is to infer an unknown target function $f(\cdot): \mathbb{R}^m \!\rightarrow\! \mathbb{R}$ using a estimated function $\hat{f}(\cdot)$ based on given an sequential stream of input-output data pairs $(\boldsymbol{x}^s, y^s)$ indexed by $s\to \infty$, where $m,s\in \mathbb{N}_{>0}$. 
Each input $\bm{x}^s \in \mathbb{R}^m$ is paired with an output $y^s = f(\boldsymbol{x}^s) + \varepsilon \in \mathbb{R}$, where the noise \(\varepsilon\) follows a normal distribution with zero mean and variance \(\sigma_n^2\), and \(\sigma_n \in \mathbb{R}_{> 0}\). 

Gaussian Processes represent a Bayesian non-parametric approach commonly employed for function regression and approximation tasks. 
It assumes that the target function \(f(\cdot)\) is sampled from a GP prior, denoted as \(f(\cdot) \sim \mathcal{GP}(m(\cdot), \kappa(\cdot, \cdot))\), where \(m(\cdot)\) is the mean function (often set to zero when no prior knowledge is assumed)~\cite{rasmussenGaussianProcessesMachine2006} and \(\kappa(\cdot, \cdot)\) is the covariance kernel (e.g., the squared exponential kernel). 
With the streaming observations accumulated up to time step \(t_k \in \mathbb{R}_{\ge 0}\) forming the dataset \(\mathbb{D}(t_k) = \{ (\boldsymbol{x}^s, y^s) \}_{s=1, 2, \dots, N(t_k)}\), where \(N(t_k) = |\mathbb{D}(t_k)|\), the posterior mean and variance functions of the GP can be derived as follows
\begin{align}
	\label{eq_mu}    
	\mu(\boldsymbol{x}) = \boldsymbol{\kappa}(\boldsymbol{x}, \boldsymbol{X}) \boldsymbol{\alpha}, &&    \sigma^2(\boldsymbol{x}) &= \kappa(\boldsymbol{x}, \boldsymbol{x}) - \boldsymbol{v}^T \boldsymbol{v},
\end{align}
where $\boldsymbol{\kappa}(\boldsymbol{x}, \boldsymbol{X}) = [\kappa(\boldsymbol{x},\boldsymbol{x}^s)]_{s=1, \dots, N(t_k)}$ is the covariance vector between the test input $\boldsymbol{x}$ and training inputs \(\boldsymbol{X}=[\boldsymbol{x}^1, \dots, \boldsymbol{x}^{N(t_k)}]\), and 
\(\boldsymbol{\alpha} \), 
$\boldsymbol{v}$ are computed using the Cholesky decomposition  $\boldsymbol{L} = \text{cholesky}(\boldsymbol{K} + \sigma_n^2 \boldsymbol{I})$ as 
\begin{align}
	\label{eq_gp_model_update}
	\boldsymbol{\alpha} = \boldsymbol{L}^{\top} \backslash (\boldsymbol{L}^{\top} \backslash \boldsymbol{y}), && \boldsymbol{v}=\boldsymbol{L} \backslash \boldsymbol{\kappa}(\boldsymbol{x}, \boldsymbol{X})
\end{align}
with the training output $\boldsymbol{y}=[y^1,\dots,y^{N(t_k)}]^{\top}$. The kernel matrix \(\boldsymbol{K}\) is defined by its elements $K_{ij}=\kappa(\boldsymbol{x}^i, \boldsymbol{x}^j)$~\cite{rasmussenGaussianProcessesMachine2006}. 

In online learning scenarios, the predictive mean and variance computations in \eqref{eq_mu} demonstrate $\mathcal{O}(N(t_k))$ and $\mathcal{O}(N^2(t_k))$ complexities respectively, indicating these operations computationally intractable as data grows without bound. 
To address these scalability challenges, a widely adopted strategy involves partitioning the data into multiple subsets and training separate GP experts with their means $\mu_i(\cdot)$ and variances $\sigma^2_i(\cdot)$, where $i\in\mathbb{N}$ denotes the index of the GP expert. 
Various aggregation techniques have been proposed for DGPs, and the structure of such aggregation methods can be formulated in a general framework as 
\begin{subequations}
	\label{eq_predictions}
	\begin{align}
		\label{eq_mu_agg}  \tilde{\mu}(\boldsymbol{x}) &= \sum\nolimits_{i\in\mathcal{N}} \omega_{i}(\boldsymbol{x}){\mu}_i(\boldsymbol{x}), \\
		\label{eq_sigma_agg}  \tilde{\sigma}^2(\boldsymbol{x}) &= \sum\nolimits_{i\in\mathcal{N}}\varpi_{i}(\boldsymbol{x}) {\sigma}_i^2(\boldsymbol{x}),
	\end{align}
\end{subequations}
where $\mathcal{N} \subset \mathbb{N}$ is the set of all GP experts, the functions $\omega_{i}(\cdot):\mathbb{R}^n \to \mathbb{R}_{\geq 0}$ and $\varpi_{i}(\cdot):\mathbb{R}^n \to \mathbb{R}_{\geq 0}$
represent the aggregation weights for the mean and variance, respectively. 

%%%%%%%%%%%%%%%%%%%%%%%%%%%%%%%%%%%%%%%%%%%%%%%%%%%%%%%%%%%%%%%%
%%%%%%%%%%%%%%%%%%%%%%%%%%%%%%%%%%%%%%%%%%%%%%%%%%%%%%%%%%%%%%%%
\section{Generated Experts of Gaussian Processes}
\label{sec_SEGP}
In this section, we introduce an efficient online distributed learning framework designed to adaptively process streaming data in real time.
At the core of SkyGP is a fully online and adaptive architecture that maintains a bounded, scalable set of GP experts.
To enable dynamic expert allocation and effective aggregation, we first present the kernel-based center mechanism in \cref{subsec_kerneCenter}.
Building on this, we describe the progressive generation strategy for updating and predicting with the GP experts in \cref{subsec_SEstrategy}, followed by the aggregation strategy in \cref{subsec_agg_prediction}.
Finally, we analyze the bounded computational complexity of SkyGP in \cref{subsec_BoundedComplexity}.

%%%%%%%%%%%%%%%%%%%%%%%%%%%%%%%%%%%%%%%%%%%%%%%%%%%%%%%%%%%%%%%%
\subsection{Kernel-Induced Distance}
\label{subsec_kerneCenter}
To enable partitioning and localization for newly generated GP experts, we adopt a center-based representation of the training inputs, where the center of the $i$-th expert $\mathcal{GP}_i$ is defined as $\boldsymbol{c}_i = \frac{1}{N_i} \sum_{k=1}^{N_i} \boldsymbol{x}_i^{k}$, following~\cite{Nguyen_NeurIPS2008_Local}. 
Each expert maintains its own representative center to support fast and scalable online allocation.
To facilitate online computation, the center of each expert is updated incrementally as new data points are assigned. Specifically, when a new point \( \boldsymbol{x}^{k} \) arriving at time step \( t_k \) is allocated to the selected expert, the expert center is updated as
\begin{align}
	\label{eq_center}
	\boldsymbol{c}_i^k = {(k-1)}\boldsymbol{c}_i^{k-1}/{k} + \boldsymbol{x}^k /k.
\end{align}
To evaluate the distance between the expert center \( \boldsymbol{c}_i^k \) and the newly arrived input \( \boldsymbol{x}^k \) in the feature space induced by the kernel, we define the kernel-based distance as follows
\begin{align}
	d_i^{k} (\boldsymbol{c}_i^{k}, \boldsymbol{x}^{k}) = {1}/{\kappa( \boldsymbol{c}_i^{k}, \boldsymbol{x}^{k} )}.
\end{align}

%%%%%%%%%%%%%%%%%%%%%%%%%%%%%%%%%%%%%%%%%%%%%%%%%%%%%%%%%%%%%%%%
\subsection{Progressive Expert Generation Strategy}
\label{subsec_SEstrategy}

\subsubsection{Expert Localization and Generation}
\label{subsubse_leaau}
\begin{algorithm}[t]
	\caption{Expert Localization with an Adaptive Window}
	\label{alg:local-expert-search-insert}
	\begin{algorithmic}[1]
		\Require $\boldsymbol{x}^k$, $\nu_{\text{prev}}$, $\{ \vartheta_{i}^{k-1} \}_{i \in \mathcal{N}(t_{k-1})}$
		
		\Statex \textbf{Step \Romannum{1}: Compute adaptive window size}
		
		\State \textbf{if} $k > 1$ \textbf{then}
		\State ~~~~~~Compute distance: $d_{\text{temp}} \gets 1/\kappa(\boldsymbol{x}^{k-1}, \boldsymbol{x}^{k}) $
		\State ~~~~~~Window size: $W \gets \min(\bar{W},\ \lfloor \exp(d_{\text{temp}} /\varrho) \rfloor )$
		\State \textbf{else} $W \gets 0$ \textbf{end if}
		
		\Statex \textbf{Step \Romannum{2}: Locate nearest expert}
		
		\State \textbf{if} $W \ne 0$ \textbf{then}
		\State ~~~~~~$\mathcal{I} \gets \{ \nu_{\text{prev}} - W, \ldots, \nu_{\text{prev}} + W \}$ from $\mathcal{T}(t_{k-1})$
		\State ~~~~~~$\mathcal{I} \gets \{ i \in \mathcal{I} \cap \mathcal{N}(t_{k-1}) \mid \vartheta_i^{k-1} > \bar{\vartheta} \}$
		\State \textbf{else} $\mathcal{I} \gets \nu_{\text{prev}}$ \textbf{end if}
		
		\State $\nu_{\text{nr}} \gets \arg\max_{i \in \mathcal{I}} \kappa(\boldsymbol{x}^k, \boldsymbol{c}_i)$; $\nu_{\text{prev}} \gets \nu_{\text{nr}}$
		\State $\mathcal{GP}_{\text{nr}}(t_k) \gets$ Choose GP expert corresponding to $\nu_{\text{nr}}$
		
		\State \Return $\mathcal{I}$ and $\nu_{\text{prev}}$
		
	\end{algorithmic}
\end{algorithm}

\begin{figure}[t]
	\centering
	\includegraphics[width=1.0\linewidth]{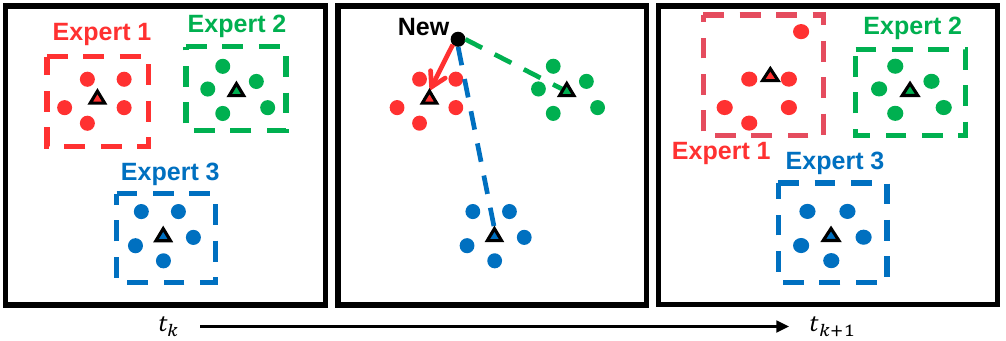}
	\caption{Data allocation process}
	\label{fig_data_process}
\end{figure}

In conventional online GP approximation methods, new data points are typically assigned to the first available expert that has not yet reached its capacity~\cite{Nguyen_NeurIPS2008_Local, lederer_ICML2021_Gaussian,yang_AAAI2025_Asynchronous}. 
Once the expert becomes full, it is either no longer updated or directly split into sub-experts, often without explicitly accounting for the underlying distributional properties and space features of the data. 
In contrast, our SkyGP framework dynamically allocates each incoming data point to the most appropriate expert based on kernel proximity and time-aware factor. 
Specifically, from the prediction phase, we have already calculated the kernel distances between the new arriving point and the experts within an adaptive window $W \in \mathbb{I}_{[1,\bar{W}]}$ and found the nearest experts with in the expert set $\mathcal{N}(t_k)$, where $\mathcal{N}(t_0)=1$ (see \cref{alg:local-expert-search-insert}). 
After locating the previous index $\nu_{\text{prev}}$ of the nearest expert denoted by $\nu_{\text{nr}}$, we are able to process the streaming data based on the index list $\mathcal{I}$ filtered by the time-aware factor $\vartheta \in (0,1]$ initialized by $1$, which encodes the usage history of the experts. 
Moreover, we define the upper bound of $\vartheta$ as $\bar{\vartheta}$ to filter out the outdated experts.
If the expert has not reached its capacity, the data point is added to it (see \cref{fig_data_process}). 
Otherwise, the two SkyGP variants handle this situation in two different ways outlined in \cref{alg_predict_update}.

In particular, a data replacement strategy is employed in SkyGP-Dense. 
When the new data pair $(\boldsymbol{x}^k, y^k)$ is confirmed to add to $\mathbb{D}_{\text{nr}}(t_k)$, and the data point within $\mathcal{GP}_{\text{nr}}$ that is furthest from its center is cast off and moved to a separate dropped dataset $\mathbb{D}_{\text{nr}}^{\text{off}}(t_k)$, whose center $^{\text{off}}\!\boldsymbol{c}_{\text{nr}}^k$ is then updated in the same method in \eqref{eq_center}. 
Since such a replacement requires recomputing the expert's kernel matrix and Cholesky factorization, we introduce an event-triggered mechanism to limit updates only to critical cases.
Let $\boldsymbol{c}_{\text{nr}}$ and $^{\text{off}}\!\boldsymbol{c}_a$ be the center and the dropped center of expert $\mathcal{GP}_{\text{nr}}$ with $N_{\text{nr}}(t_k)$ data points $\{ (\boldsymbol{x}_{\text{nr}}^{s} ,y_{\text{nr}}^s\}_{s=1}^{N_{\text{nr}}(t_k)}$. 
The event-triggered data forwarding strategy is designed as
\begin{align}
	\label{eq_D_nr_tk}
	\mathbb{D}_{\text{nr}}(t_k) \!\!=\!\! \begin{cases}
		\!\mathbb{D}_{\text{nr}}(t_{k\!-\!1}) \!\!\cup\!\! (\boldsymbol{x}^k,\! y^k) \backslash (\boldsymbol{x}^{k_\text{off}},\! y^{k_\text{off}})\!,\!\! & \! \text{ if }  \Delta < 0 \\
		\!\mathbb{D}_{\text{nr}}(t_{k\!-\!1}),\!\! & \!\text{ otherwise } 
	\end{cases}\!\!\!, 
\end{align}
where $\Delta \in \mathbb{R}$ is defined as $\Delta = \max_{s = 1, \cdots, N_{\text{nr}}(t_k)} \Delta(s)$ and
\begin{align}
	\Delta(s) \!=\! \kappa(\boldsymbol{x}^{s}, \boldsymbol{c}_{\text{nr}}) \!-\! \kappa(\boldsymbol{x}^s, ^{\text{off}}\!\boldsymbol{c}_{\text{nr}}) \!-\! \kappa(\boldsymbol{x}^{k}, \boldsymbol{c}_{\text{nr}}) \!+\! \kappa(\boldsymbol{x}^k, ^{\text{off}}\!\boldsymbol{c}_{\text{nr}}). \nonumber
\end{align}
This kernel-based selection strategy encourages experts to retain points that align with the current local distribution while avoiding repeated inclusion of previously discarded regions. 
Although computationally expensive, this step is infrequent and remains tractable under the predefined data threshold $\bar{N}$.
If no expert accepts \( \boldsymbol{x}^{k} \), a new expert will be created and the center will be initialized with \( \boldsymbol{x}^{k} \).
In {SkyGP-Fast}, no replacement is performed. 
This design ensures all update GP expert is a rank-one update to the Cholesky factor, which incurs a computational complexity of \( \mathcal{O}(N_{\text{nr}}^2) \).
{SkyGP-Fast} bypasses recomputations by appending new experts instead of replacing data, favoring low-latency updates and scalability. 
In contrast, {SkyGP-Dense} limits memory usage by reusing expert slots, making it more suitable for resource-constrained settings. 
Together, these two modes offer a flexible trade-off between computational efficiency and memory control across diverse online learning scenarios.

\begin{algorithm}[!t]
	\caption{Expert Update Strategy}
	\label{alg_predict_update}
	\begin{algorithmic}[1]
		\Require $\boldsymbol{x}^k$, $\nu_{\text{prev}}$,  $\mathcal{I}$, $\{ \vartheta_{i}^{k-1} \}_{i \in \mathcal{N}(t_{k-1})}$

		\Statex \textbf{Step \Romannum{1}: Aggregated Prediction}
		\State $\mathcal{I}_{\text{temp}} \gets \{ \nu \in \mathcal{I} \mid \vartheta_{\nu}^{k-1} > \bar{\vartheta} \}$
        
		\State $\mathcal{I}_{\text{agg}} \gets$
		indices of the largest $\bar{\mathcal{N}}$ values in $\mathcal{I}_{\text{temp}}$
		
		\State $\tilde{\mu}(\boldsymbol{x}^k), \tilde{\sigma}(\boldsymbol{x}^k) \gets$ \eqref{eq_predictions} with MOE/POE/BCM~\eqref{eq_moe} to \eqref{eq_agg_bcm}

		\Statex \textbf{Step \Romannum{2}: GP Expert Model Update}
		
		\State $\{ \vartheta_{i}^{k} \}_{i \in \mathcal{N}} \gets \{ \vartheta_{i}^{k-1} \}_{i \in \mathcal{N}}$
		
		\ForAll{$\nu \in \mathcal{I}_{\text{agg}}$}
		
		\State $\vartheta_{\nu}^k \!\!\gets\!\! 1$, \!$\mathbb{D}_{\text{temp}} \!\!\gets\! \{ \boldsymbol{x} \!\!\in\! \mathbb{D}_{\nu}(t_{k-1})| \kappa(\boldsymbol{x}, 
		\boldsymbol{c}_{\nu}) \!\!>\!\! \kappa(\boldsymbol{x}^k, 
		\boldsymbol{c}_{\nu}) \}$
		
		\If{$|\mathbb{D}_{\nu} (t_{k-1})| < \bar{N}$}
		\State $ \mathbb{D}_{\nu}(t_{k}) \gets \mathbb{D}_{\nu}(t_{k-1}) \cup (\boldsymbol{x}^k, y^k)$ 
		\State Update $\mathcal{GP}_{\nu}(t_{k}) \gets$ Cholesky rank-one update
		\State \Return $\{ \vartheta_{i}^{k} \}_{i \in \mathcal{N}(t_k)}$
		\ElsIf{\texttt{SkyGP-Dense} and $\mathbb{D}_{\text{temp}} = \emptyset$}
		
		\State Update $\mathbb{D}_{\nu}(t_{k}) \gets$ \eqref{eq_D_nr_tk}, $\mathcal{GP}_{\nu}(t_{k})$  $\gets$ \eqref{eq_gp_model_update}
		\State Update $^{\text{off}}\boldsymbol{c}_{\nu}\gets$ \eqref{eq_center} with dropped data
		\State \Return $\{ \vartheta_{i}^{k} \}_{i \in \mathcal{N}(t_k)}$
		
		\EndIf
		\EndFor
		
		\State Create $\mathcal{GP}_{|\mathcal{N}(t_{k-1})|+1}$ with dataset $\mathbb{D}_{}(t_k)=(x_t, y_t)$
		% \State Update expert list with $\mathcal{N}(t_k) \gets \{ \mathcal{N}(t_{k-1}), | \mathcal{N}(t_{k-1}) | + 1\}$ 
		\State Update expert list $\mathcal{T}(t_k) \gets \mathcal{T}(t_{k-1})$ and \eqref{eqn_nu_new}
		\State Update $\mathcal{N}(t_k) \gets \{ \mathcal{N}(t_{k-1}), | \mathcal{N}(t_{k-1}) | + 1\}$ 
		
		\Statex \textbf{Step \Romannum{3}: Update Time-Aware Factor}
		
		\State \textbf{for} $i = \mathcal{N}\backslash\mathcal{I}_{\text{agg}}$ ~\textbf{do} $\vartheta_{i}^{k} \gets \rho \vartheta_{i}^{k-1}$ \textbf{end for}
		
		\State \Return $\{ \vartheta_{i}^{k} \}_{i \in \mathcal{N}(t_k)}$
		
	\end{algorithmic}
\end{algorithm}

\subsubsection{Dynamic Expert List Creation}
\label{local_expert_search}
The expert search is implemented over a center-indexed list that maintains a globally ordered list of expert centers based on their insertion positions. 
The index of this list is defined as $\nu_i \in \mathbb{N}$ with corresponding $\mathcal{GP}_i$ expert. 
The newly instantiated expert with center \( \boldsymbol{x}^{k} \) will be inserted adjacent to the nearest existing expert identified during search.
To update the table when new GP expert generated, we compare the kernel-based distance between \( \boldsymbol{x}^{k} \) and the left/right neighbors of the selected experts with their corresponding center $\boldsymbol{c}_{\text{nr}}^{-}$ and $\boldsymbol{c}_{\text{nr}}^{+}$:
\begin{align}
	d_{\text{left}} = 1/\kappa(\boldsymbol{x}^{k}, \boldsymbol{c}_{\text{nr}}^{-}), && \quad d_{\text{right}} = 1/\kappa(\boldsymbol{x}^{k} - \boldsymbol{c}_{\text{nr}}^{+})
\end{align}
According to the previous $\nu_{nr}$ in \cref{alg:local-expert-search-insert}, the new index of $\mathcal{GP}_{new}$ is 
\begin{align} \label{eqn_nu_new}
	\nu_{\text{new}} = 
	\begin{cases}
		\min(\nu_{\text{nr}}\ |\mathcal{N}|), & \text{if } d_{\text{right}} < d_{\text{left}} \\
		\max(\nu_{\text{nr}}-1,\ 0)  & \text{otherwise}
	\end{cases}.
\end{align}

\subsection{Aggregated Prediction}
\label{subsec_agg_prediction}
Leveraging DGP framework in \eqref{eq_predictions}, our proposed SkyGP can be seamlessly integrated with existing aggregation strategies to enhance predictive performance and uncertainty estimation across decentralized data streams. 
These weights are typically chosen to satisfy certain properties, such as non-negativity and normalization, ensuring that the aggregated predictions are coherent and interpretable.
For example, the aggregated weight functions of MoE are 
\begin{subequations}
	\label{eq_moe}
	\begin{align}
		\label{eq_agg_mean_moe} \omega_{i}(\boldsymbol{x}) &= w_i, \\
		\label{eq_agg_variance_moe} \varpi_{i}(\boldsymbol{x}) &= w_i\big(\sigma_i^2(\boldsymbol{x}) + \mu_i^2(\boldsymbol{x}) \big) -\tilde{\mu}^2(\boldsymbol{x})/w_i,
	\end{align}
\end{subequations}
satisfying $\sum_{i\in\mathcal{I}_{\text{agg}}} w_i =1$, which is also utilized in the mixture of explicitly localized experts. 
Considering the posterior variance in GP experts, the PoE family methods employ the following aggregation structure
\begin{subequations}
	\label{eq_poe}
	\begin{align}
		\label{eq_agg_mean_poe} \omega_{i}(\boldsymbol{x}) &= w_i \sigma_i^2(\boldsymbol{x}) / \varpi_{i}(\boldsymbol{x}), \\
		\label{eq_agg_variance_poe} \varpi_{i}(\boldsymbol{x}) &= 1/\sum\nolimits_{s\in\mathcal{I}_{\text{agg}}} w_s \sigma_s^{-2}(\boldsymbol{x}).
	\end{align}
\end{subequations}
Furthermore, these aggregation schemes can be extended to incorporate the prior variance of the unknown function, denoted as $\sigma_{*}$, to further refine the combined predictive uncertainty in the BCM family approaches. 
In this context, the weight function for the mean remains consistent with that of the PoE in \eqref{eq_agg_mean_poe}, while the corresponding weight function for the aggregated posterior variance is given by
\begin{align}
	\label{eq_agg_bcm} \varpi_{i}(\boldsymbol{x}) \!\!=\!\! 1 \!/\! \Big(\sum\nolimits_{s\in\mathcal{I}_{\text{agg}}} \!\!\! w_j \sigma_j^{-2}(\boldsymbol{x}) \!\!+\!\! \big( 1 \!\!-\!\! \sum\nolimits_{j\in\mathcal{I}_{\text{agg}}} \!\!\! w_j \!\big) \sigma_*^{-2}\Big)\!.\!
\end{align}
This extension ensures that the aggregated predictions remain well-calibrated, especially in distributed or federated learning scenarios where each expert may have access to different subsets of the data.

According to \cref{alg:local-expert-search-insert}, given a new input \( \boldsymbol{x}^k \), SkyGP performs prediction by first identifying the definable \(\bar{\mathcal{N}}\) number of experts within the search window. 
Based on the returned aggregation expert list $\mathcal{I}_{\text{agg}}$, each expert predicts the posterior mean \( \mu_i(\boldsymbol{x}^k) \) and variance \( \sigma_i^2(\boldsymbol{x}^k) \). 
These predictions then are fused using a principled weighting strategy, e.g., MoE, PoE or BCM, yielding the final prediction in \eqref{eq_predictions}.

%%%%%%%%%%%%%%%%%%%%%%%%%%%%%%%%%%%%%%%%%%%%%%%%%%%%%%%%%%%%%%%%
\subsection{Bounded Complexity Analysis}
\label{subsec_BoundedComplexity}
In this section, we formalize the computational efficiency of the proposed {SkyGP} framework, which is designed to maintain bounded per-step complexity during both model update and prediction. 
The primary sources of computational cost include nearest expert localization, aggregated prediction, and expert model update. 
Each component is explicitly designed to ensure low complexity and scalability in streaming scenarios, while maintaining high prediction accuracy.

\subsubsection{Model Update}
For expert localization, as described in Algorithm~\ref{alg:local-expert-search-insert}, the computational cost from evaluating kernel similarities between the current input and the expert centers to obtaining index set $\mathcal{I}$ (line 1-8) is $\mathcal{O}(1)$. 
The subsequent selection of the nearest expert from the candidate set $\mathcal{I}$ in line 9, even when incorporating the time-aware factor, requires only a linear search over $\mathcal{I}$ with the same order of complexity, i.e., $\mathcal{O}(W)$. Consequently, the overall computational complexity of Algorithm~\ref{alg:local-expert-search-insert} is $\mathcal{O}(W)$.

The overall computational cost of the \cref{alg_predict_update} is dominated by three components. 
First, the selection of the aggregated expert set in line 2 requires filtering the candidate index set $\mathcal{I}$ and selecting the top $\bar{\mathcal{N}}$ elements, which can be performed in $\mathcal{O}(|\mathcal{I}|\log \bar{\mathcal{N}})$ time. 
Second, the update phase iterates over the $\bar{\mathcal{N}}$ aggregated experts.
For each, the construction of the temporary data subset incurs $\mathcal{O}(|\mathbb{D}_{\nu}(t_{k-1})|)$ cost, which is bounded by $\mathcal{O}(\bar{N})$ considering $|\mathbb{D}_{\nu}(t_k)| \le \bar{N}$.
Moreover, when a GP model is updated via a rank-one Cholesky modification for {SkyGP-Fast}, an additional $\mathcal{O}(\bar{N}^2)$ cost per expert is incurred. 
On the other hand, {SkyGP-Dense} requires full Cholesky factor recomputation when performing replacement operations, incurring a higher cost of \( \mathcal{O}(\bar{N}^3) \). 
Third, the time-aware weighting update over the remaining experts requires $\mathcal{O}(\bar{\mathcal{N}})$ operations. 
Hence, the total worst-case complexity can be expressed as $\mathcal{O}(W \log \bar{\mathcal{N}} + \bar{\mathcal{N}}\bar{N}^2)$ for {SkyGP-Fast}, and similarly $\mathcal{O}(W \log \bar{\mathcal{N}} + \bar{\mathcal{N}}\bar{N}^3)$ for {SkyGP-Dense}. 

\subsubsection{Prediction}
Aggregated mean and variance in {SkyGP} are computed using the bounded top \( \bar{\mathcal{N}} \) nearest experts. 
As each local GP performs inference with \( \mathcal{O}(N^2(t_k)) \) complexity, the overall prediction complexity per step is \( \mathcal{O}( \bar{\mathcal{N}} N^2(t_k)) \).

\begin{remark}
	Other auxiliary operations in {SkyGP}, such as list insertion, temporal decay updates, incur at most linear complexity with respect to the number of queried experts \( \bar{\mathcal{N}} \), i.e., \( \mathcal{O}(\bar{\mathcal{N}}) \). As these operations are not computationally dominant, we omit their detailed analysis for clarity.
\end{remark}

%%%%%%%%%%%%%%%%%%%%%%%%%%%%%%%%%%%%%%%%%%%%%%%%%%%%%%%%%%%%%%%%
%%%%%%%%%%%%%%%%%%%%%%%%%%%%%%%%%%%%%%%%%%%%%%%%%%%%%%%%%%%%%%%%
\section{Safe Learning-based Control Policy}
In this section, we present a learning-based control policy that integrates the proposed SkyGP framework into nonlinear systems with unknown dynamics.
The key idea is to leverage real-time GP predictions to estimate model uncertainties and design a feedback controller that ensures safe and stable trajectory tracking.
Following the introduction of the prediction error bound of SkyGP in \cref{subsec_error_bound}, we analyze the control performance for general dynamical systems and design a tracking control policy specifically for Euler–Lagrange (EL) systems in \cref{subsec_safe_control}.

\subsection{Learning Performance of SkyGP}
\label{subsec_error_bound}

We consider a function \( f \) belonging to a reproducing kernel Hilbert space (RKHS) \( \mathcal{H}_\kappa \) associated with a positive definite kernel \( \kappa(\cdot, \cdot) \), such that \( \|f\|_\kappa \leq \Gamma \). 
Let \( \mu_i(\bm{x}) \) and \( \sigma_i(\bm{x}) \) denote the posterior mean and standard deviation from the $i$-th GP expert trained on dataset \( \mathbb{D}_i \) for $i = 1, \cdots, \mathcal{N}$. 
Then the following probabilistic prediction error bound holds.

\begin{lemma}
	\label{lemma:gp-error-bound}
	Consider the regression task in a compact domain $\mathbb{X}$, and suppose the kernel function $\kappa(\cdot, \cdot)$ is Lipschitz with Lipschitz constant $L_{\kappa} \in \mathbb{R}_{0,+}$.
	Choose $\delta \in (0,1 / \bar{\mathcal{N}})$ and $\tau \in \mathbb{R}_+$, then the prediction error satisfies
	\begin{align}
		|f(x) - \tilde{\mu}(\bm{x} \mid \mathbb{D})| \leq \beta \sigma(\bm{x}) + \gamma(\bm{x}), && \forall \bm{x} \in \mathbb{X}
	\end{align}
	with a probability of at least $1 - \bar{\mathcal{N}} \delta$, where $\gamma(\bm{x}) = \sum\nolimits_{i \in \mathcal{N}} \omega_i(\bm{x}) \gamma_i$, $\sigma(\bm{x}) = \sum\nolimits_{i \in \mathcal{N}} \omega_i(\bm{x}) \sigma_i(\bm{x})$,  and
	\begin{align}
		&\beta = 2 \Big( 2 \sum\nolimits_{j=1}^{n} \log{\Big \lceil \frac{\sqrt{n}} {2 \tau} (\bar{x}_{j} - \underline{x}_{j}) \Big \rceil } - 2\log{ \frac{\delta}{n}} \Big)^{1/2}, \\
		&\gamma_i = \big( \sqrt{\beta_{\delta}} L_{\sigma,i} + \Gamma \sqrt{2 L_{\kappa}} + L_{\mu,i} \big) \tau,
	\end{align}
	with $\bar{x}_j = \max_{x \in \mathbb{X}} x_j$, $\underline{x}_j = \min_{x \in \mathbb{X}} x_j$ and $x_j$ as the $j$-th dimension of $x$.
	The positive constants $L_{\mu,i} \in \mathbb{R}_{0,+}$ and $L_{\sigma,i} \in \mathbb{R}_{0,+}$ are the Lipschitz constant for posterior mean $\mu_i(\cdot)$ and variance $\sigma_i(\cdot)$, respectively for $i \in \mathcal{N}$.
\end{lemma}

\cref{lemma:gp-error-bound} shows the probabilistic theoretical error bound of the aggregated prediction $\tilde{\mu}(\cdot)$, which also holds for the proposed Sky-GP\footnote{The proofs of all lemmas and theorems are provided in Appendix A.}.
Note that the boundness of the prediction error does not rely on the choice of aggregating strategy reflected by the weighting function $\omega_i(\cdot)$.
For notational simplicity, denote the prediction error bound as $\eta(\bm{x}) = \beta \sigma(\bm{x}) + \gamma(\boldsymbol{x})$, which is used in the following control analysis.

\begin{remark}
	Compared with the other probabilistic \cite{srinivas2012information,scharnhorst2022robust} or deterministic bounds \cite{maddalena2021deterministic,hashimoto2022learning}, the error bound chosen in this paper follows \cite{lederer2019uniform}, which results in an amplitude constant $\beta$ of posterior variance $\sigma(\cdot)$.
	Moreover, it is shown that there always exists a sufficiently small grid factor $\tau \in \mathbb{R}_+$ such that $\gamma(\bm{x}) \ll \beta \sigma(\bm{x})$ for all $\bm{x} \in \mathbb{X}$, i.e., the posterior variance dominates the error bound $\eta(\bm{x})$.
	Therefore, the prediction error bound could also be written as $\eta(\bm{x}) \le 2 \beta \sigma(\bm{x})$ \cite{lederer2024safe}.
\end{remark} 

%%%%%%%%%%%%%%%%%%%%%%%%%%%%%%%%%%%%%%%%%%%%%%%%%%%%%%%%%%%%%%%%
\subsection{Safe Critical Control Application}
\label{subsec_safe_control}
\subsubsection{A General Control System}
The objective is to design a GP-based control law that drives an unknown dynamical system to exhibit a desired behavior. 
Specifically, we consider a class of control applications focused on output stabilization, where the goal is to ensure that the system output converges to the equilibrium point asymptotically.
The continuous-time nonlinear system is described as follows
\begin{align}
	\label{eq_system}
	\dot{\bm{x}} = \bm{f}(\bm{x}, \bm{u}) + \bm{g}(\bm{x}, \bm{u}), &&
	\bm{z} = \bm{h}(t, \bm{x})
\end{align}
where $\bm{x} \in \mathbb{X} \subset \mathbb{R}^{m}$, $\bm{u} \in \mathbb{U} \subset \mathbb{R}^{m_u}$ and $\bm{z} \in \mathbb{R}^{m_z}$ denote the system state, control input and system output, respectively.
The transition function $\bm{f}(\cdot,\cdot) = [f_1(\cdot,\cdot), \cdots, f_{m}(\cdot,\cdot)]^{\top}: \mathbb{X} \times \mathbb{U} \to \mathbb{R}^{m}$ is unknown, where each scalar $f_j(\cdot,\cdot)$ belongs to a RKHS $\mathcal{H}_{\kappa,j}$ corresponding to a kernel function $\kappa_j(\cdot, \cdot): \mathbb{X}_e \times \mathbb{X}_e \to \mathbb{R}_{0,+}$ with $\mathbb{X}_e = \mathbb{X} \times \mathbb{U} \subset \mathbb{R}^{m + m_u}$ for all $j = 1, \cdots, m$. 
The measurement function $\bm{h}(\cdot, \cdot): \mathbb{R}_{\ge 0} \times \mathbb{X} \to \mathbb{R}^{m_z}$ is known.
To achieve system output stabilization, i.e., $\lim_{t \to \infty} \bm{z}(t) = \bm{0}_{m_z \times 1}$, a control policy $\bm{\pi}(\cdot, \cdot, \cdot): \mathbb{R}_{\ge 0} \times \mathbb{X} \times \mathcal{H}_{\kappa} \to \mathbb{R}^{m_u}$ is designed satisfying the following assumption.
\begin{assumption}
	\label{ass_controller}
	Suppose there exists a differentiable function $V(\cdot, \cdot): \mathbb{R}_{\ge 0} \times \mathbb{R}^{m_y} \to \mathbb{R}_{\ge 0}$ satisfying
	\begin{align}
		\underline{\alpha}(\| \bm{z} \|) \le V(t, \bm{z}) \le \bar{\alpha}(\| \bm{z} \|)
	\end{align}
	with class-$\mathcal{K}$ functions $\underline{\alpha}(\cdot), \bar{\alpha}(\cdot): \mathbb{R}_{\ge 0} \to \mathbb{R}_{0,+}$.
	The control law $\bm{\pi}(\cdot, \cdot, \cdot)$ is designed such that
	\begin{align} \label{eqn_ISS_dotV_condition}
		& \nabla_{\bm{x}} V(t, \bm{x}) (\bm{\mu}(\bm{x}, \bm{\pi}(t, \bm{x}, \bm{\mu})) + \bm{g}(\bm{x}, \bm{\pi}(t, \bm{x}, \bm{\mu})))  \\
		&+\nabla_t V(t,\bm{x})\le - \alpha(V(t, \bm{h}(t, \bm{x}))) - \| \nabla_{\bm{x}} V(t, \bm{x}) \|^2 / (4 \varepsilon), \nonumber
	\end{align}
	with a positive constant $\varepsilon \in \mathbb{R}_{>0}$, where the function $\alpha(\cdot): \mathbb{R}_{\ge 0} \to \mathbb{R}_{\ge 0}$ belongs to class-$\mathcal{K}$ and
	\begin{align}
		\nabla_t V(t, \bm{x}) &= \frac{\partial V(t, \bm{h}(t, \bm{x}))}{\partial t} + \frac{\partial V(t, \bm{h}(t, \bm{x}))}{\partial \bm{h}(t, \bm{x})} \frac{\partial \bm{h}(t, \bm{x})}{\partial t}, \nonumber \\
		\nabla_{\bm{x}} V(t, \bm{x}) &= \frac{\partial V(t, \bm{h}(t, \bm{x}))}{\partial \bm{h}(t, \bm{x})} \frac{\partial \bm{h}(t, \bm{x})}{\partial \bm{x}},
	\end{align}
	for any $\bm{\mu}(\cdot) = [\mu_1(\cdot), \cdots, \mu_{m}(\cdot)]^T$ satisfying $\mu_i(\cdot) \in \mathcal{H}_{\kappa,j}$ for all $j = 1, \cdots, m$.
\end{assumption}
This assumption indicates the asymptotic output stability of the equivalent system $\dot{\bm{x}} = \bm{\mu}(\bm{x}, \bm{u}) + \bm{g}(\bm{x}, \bm{u})$ using the control law $\bm{\pi}(t, \bm{x}, \bm{\mu})$.
Compared to the condition in conventional asymptotic stability \cite{khalil2015nonlinear}, the additional term $\| \nabla_{\bm{x}} V(t, \bm{x}) \|^2 / (4 \varepsilon)$ introduced in \eqref{eqn_ISS_dotV_condition} is used following input-to-state control Lyapunov functions to partially compensate for the potential effect from unknown disturbances. 
Later, an example is provided to demonstrate how to design a model-based controller that satisfies this assumption.
Notably, the true dynamical system is unknown $\bm{f}(\cdot)$ instead of $\bm{\mu}(\cdot)$, whose performance is shown as follows. 

\begin{theorem}
	\label{theorem_boundZ}
	If there exists a control law $\boldsymbol{\pi}(\cdot, \cdot, \cdot)$ satisfying Assumption~\ref{ass_controller}.
	Choose $\delta \in (0,1/m)$, then the output $\boldsymbol{z}(t)$ of the true system~\eqref{eq_system} is ultimately bounded  by
	\begin{align}
		\lim\nolimits_{t \to \infty} \| \bm{z}(t) \| \le \underline{\alpha}^{-1}(\alpha^{-1}(\varepsilon \bar{\eta}^2))
	\end{align}
	with a probability of at least $1 - m \delta$.
	The positive constant $\bar{\eta} = \max_{t \in \mathbb{R}_{0,+}, \bm{x} \in \mathbb{X}} \| \bm{\eta}(\bm{\xi}(t, \bm{x}, \bm{\mu})) \|$ denotes an upper bound on the model error between $\boldsymbol{f}(\cdot)$ and its GP approximation $\boldsymbol{\mu}$, where $\bm{\xi}(t, \bm{x}, \bm{\mu}) = [\bm{x}^T, \bm{\pi}^T(t, \bm{x}, \bm{\mu})]^T$ and $\bm{\eta}(\cdot) = [\eta_1(\cdot), \cdots, \eta_{d_x}(\cdot)]^T$ and $\eta_i(\cdot)$ obtained using prediction error bound.
\end{theorem}
This theorem guarantees that the system output ultimately remains within a bounded neighborhood of the desired equilibrium, despite the presence of model uncertainties. 
To instantiate this theorem and derive a practically applicable controller, we next consider a representative Euler–Lagrange (EL) system and demonstrate the design and performance of a corresponding learning-based policy.

\subsubsection{Euler-Lagrange System Control}

Consider an EL system
\begin{align}
	\label{eq_ELsystem}
	\bm{M}(\bm{q}) \ddot{\bm{q}} + \bm{C}(\bm{q}, \dot{\bm{q}}) \dot{\bm{q}} + \bm{g}(\bm{q}, \dot{\bm{q}}) = \bm{u} + \bm{d}(\bm{q}, \dot{\bm{q}}),
\end{align}
where $\bm{q}\! \in\! \mathbb{Q} \!\subset\!\mathbb{R}^{m_q}$ and $\bm{u} \! \in \! \mathbb{U} \!\subset \!\mathbb{R}^{m_u}$ denote the generalized coordinate and control input with $m_q = m / 2$.
Define $\bm{x} = [\bm{q}^{\top}, \dot{\bm{q}}^{\top}]^{\top} \in \mathbb{X}$, then the dynamics of \eqref{eq_ELsystem} is rewritten as
\begin{align}
	\dot{\bm{x}} \!=\! \underbrace{\begin{bmatrix}
			\dot{\bm{q}} \\
			\bm{M}^{-1}(\bm{q}) (\bm{u} - \bm{C}(\bm{q},\dot{\bm{q}}) \dot{\bm{q}} - \bm{g}(\bm{x}))
	\end{bmatrix}}_{\bm{g}(\bm{x}, \bm{u})} \!+\! \underbrace{\begin{bmatrix}
			\bm{0}_{m_q \times 1} \\
			\bm{M}^{-1}(\bm{q}) \bm{d}(\bm{x})
	\end{bmatrix}}_{\bm{f}(\bm{x},\bm{u})} \nonumber 
\end{align}
The control task is to track a pre-defined trajectory $\bm{x}_d(\cdot) = [\bm{q}_d^{\top}(\cdot), \dot{\bm{q}}_d^{\top}(\cdot)]^{\top}: \mathbb{R}_{\ge 0} \to \mathbb{X}$, such that the output is defined as tracking error as $\bm{z}(t) = \bm{x}(t) - \bm{x}_d(t)$.
To achieve this control objective, a learning-based control law inspired by computed torque control is proposed as
\begin{align}
	\label{eq_ELcontroller}
	&\bm{\pi}(t, \bm{x}, \bm{\mu}) = \bm{C}(\bm{x}) \dot{\bm{q}} + \bm{g}(\bm{x}) - \bm{\mu}(\bm{x}) + \bm{M}(\bm{q}) \ddot{\bm{q}}_d(t) \\
	&+\!\! \bm{M}(\bm{q}) (\bm{K}_p (\bm{q} \!\!-\!\! \bm{q}_d(t)) \!\!+ \!\! \bm{K}_d (\dot{\bm{q}} \!-\! \dot{\bm{q}}_d(t)) ) \!\!-\!\! \bm{B}^T\!(\bm{x}) \bm{P} \bm{z} \!/\! (2 \varepsilon),\! \nonumber
\end{align}
where $\boldsymbol{B}= [
\bm{0}_{d_q \times d_q} , \bm{M}^{-\top}(\bm{q}) ]^{\top}$. The control gains $\bm{K}_p$ and $\bm{K}_d$ are chosen such that 
$\bm{A}=\begin{bmatrix}
	\bm{0}_{m_q \times m_q} & \bm{I}_{m_q} \\
	\bm{K}_p & \bm{K}_d
\end{bmatrix}$ is Hurwitz.
The matrix $\bm{P}$ is the solution to the Lyapunov equation associated with the quadratic Lyapunov function $V(t, \bm{z}) = \bm{z}^\top \bm{P} \bm{z}$, where $\bm{P} \succ 0$ satisfies $
\bm{A}^\top \bm{P} + \bm{P} \bm{A}  = - \bm{Q}$ with $\bm{Q} \succ 0$ being a positive definite matrix.
The existence and uniqueness of the solution $\bm{P}$ is guaranteed due to Hurwitz $\bm{A}$ \cite{khalil2015nonlinear}.
Then, the control performance in terms of tracking error bound is shown as follows.

\begin{theorem}
	\label{theorem_tracking}
	Consider the EL system \eqref{eq_ELsystem} driven by the control law \eqref{eq_ELcontroller} using the proposed Sky-GP satisfying all assumptions in \cref{lemma:gp-error-bound}.
	Choose $\delta \in (0, 1 / m_q) \subset \mathbb{R}$, then the tracking error $\bm{z}$ is ultimately bounded by
	\begin{align}
		\lim\nolimits_{t \to \infty} \| \bm{z}(t) \| \le \varepsilon \bar{\lambda}(\bm{P}) (\underline{\lambda}(\bm{Q})  \underline{\lambda}(\bm{P}) )^{-1} \bar{\eta}^2
	\end{align}
	with a probability of at least $1 - m_q \delta$.
\end{theorem}

\cref{theorem_tracking} shows the tracking error bound obtained by using the controller \eqref{eq_ELcontroller} with the proposed Sky-GP, which is relevant to the worst-case learning performance $\bar{\eta}$ and the desired convergence strength reflected by $\bm{Q}$.
Specifically, smaller prediction error $\bar{\eta}$ and larger eigenvalues of $\bm{Q}$ induce smaller tracking error.
Note that increasing the control gains with larger eigenvalues of $\bm{Q}$ also leads to high sensitivity of the measurement noise, which may deteriorate the control performance.
Therefore, an accurate prediction for a smaller $\bar{\eta}$ is essential to achieve high tracking precision.

%%%%%%%%%%%%%%%%%%%%%%%%%%%%%%%%%%%%%%%%%%%%%%%%%%%%%%%%%%%%%%%%
%%%%%%%%%%%%%%%%%%%%%%%%%%%%%%%%%%%%%%%%%%%%%%%%%%%%%%%%%%%%%%%%
\section{Experimental Evaluation}
\label{sec_ExperimentalEvaluation}
We conduct comprehensive experiments to evaluate the proposed SkyGP framework from both regression and control perspectives.
In \cref{subsec_regression_performance}, we assess the regression performance on multiple real-world benchmark datasets.
Then, a closed-loop control performance comparison is evaluated against SOTA approaches on a nonlinear dynamical system in \cref{subsec_control_performance}.
Finally, we present an ablation study analyzing different algorithmic design choices\footnote{For additional results and ablation study analysis, refer to Appendix B.}.

%%%%%%%%%%%%%%%%%%%%%%%%%%%%%%%%%%%%%%%%%%%%%%%%%%%%%%%%%%%%%%%%
\subsection{Regression Performance Evaluation}
\label{subsec_regression_performance}
We evaluate the proposed SkyGP framework on four real-world benchmark datasets.
The SARCOS dataset contains 44,484 samples with 21 input features.
The PUMADYN32NM (PUMA) dataset consists of 7,168 samples with 32 input features.
The KIN40K dataset includes 10,000 samples with 8-dimensional inputs.
The ELECTRIC dataset records household electricity consumption, comprising over 2 million samples with 11-dimensional inputs.
For simplicity, we use the first 20,000 samples of ELECTRIC dataset.
To assess the performance of SkyGP, we compare our two variants against several strong baselines: LoG-GP~\cite{lederer_ICML2021_Gaussian} and Local GPs~\cite{Nguyen_NeurIPS2008_Local} with each expert holding at most 50 data points, SSGP with 50 inducing points~\cite{bui_NeurIPS2017_Streaming}, and ISSGP with 200 random features~\cite {Arjan_NN2013_Real}. 
All methods use an ARD kernel with hyperparameters pre-trained with the first 1000 samples in each dataset.
Each model is evaluated in a sequential setting, where predictions and updates are performed on-the-fly as each new data point arrives. 
We evaluated the average prediction and update times, as well as the standardized mean squared error (SMSE) and the mean standardized log loss (MSLL) in a sequential interpretation~\cite{lederer_ICML2021_Gaussian}. 
The maximal number of experts \(\bar{\mathcal{N}} \in \{1, 2, 4\}\) are used in all benchmark. 
Each expert with SkyGP holds at most \(\bar{N} = 50\) data points and the maximal search window size \(\bar{W} = 40\) and uses the rBCM aggregation method. 
Set the time-aware decay weighting factor $\rho = 0.995$ and $\bar{\vartheta}= 10^{-3}$.
SkyGP-Fast with a maximum of one expert is denoted by SkyGP-F-1, and SkyGP-Dense with one expert by SkyGP-D-1. Other variants follow the same naming convention.

\cref{tab_acc} shows the average SMSE and MSLL across three benchmark datasets, demonstrating that the proposed SkyGP variants consistently outperform baseline methods such as LoG-GP and LocalGPs in both predictive accuracy and uncertainty calibration. While ISSGP achieves the best SMSE and MSLL on the PUMA dataset, its prediction and update time is over 30 times slower, as detailed in \cref{tab_time}, making it less suitable for real-time applications. 
Notably, SkyGP-D-4 achieves the best overall performance, with the lowest SMSE (0.017) and MSLL (-2.03) on the SARCOS dataset, and competitive results on PUMA and ELECTRIC. 
\cref{tab_time} summarizes the average prediction and update times across three datasets. 
SkyGP-F-1 achieves the lowest overall latency, with update times consistently at 0.04s and the fastest prediction on SARCOS (0.16s), making it well-suited for real-time applications. 
SkyGP-D-1 has the best prediction time on ELECTRIC (0.14s) while maintaining competitive update performance. 
In contrast, traditional baselines such as ISSGP and SSGP exhibit significantly higher computational costs, with ISSGP requiring up to 18s for prediction and 7s for update on SARCOS, rendering them impractical for online deployment. 
Notably, SSGP fails to produce results on the ELECTRIC dataset within 20s, so we omit it. 

\begin{table}[t]
	\centering
	\caption{Average SMSE and MSLL on 3 datasets.}
	\label{tab_acc}
	\begin{tabular}{lcccccc}
		
		\toprule
		{{Model}} 
		& \multicolumn{2}{c}{{SARCOS}} 
		& \multicolumn{2}{c}{{PUMA}} 
		& \multicolumn{2}{c}{{ELECTRIC}} \\
		\cmidrule(lr){2-3} 
		\cmidrule(lr){4-5} 
		\cmidrule(lr){6-7}
		& $\epsilon_{\text{smse}}$\! & $\epsilon_{\text{msll}}$\! 
		& $\epsilon_{\text{smse}}$\! & $\epsilon_{\text{msll}}$\! 
		& $\epsilon_{\text{smse}}$\! & $\epsilon_{\text{msll}}$\!  \\
		\midrule
		LoG-GP\!\! & \!0.044\! & -\!1.83   & 0.20 & -\!1.00 & 0.11 & \!-\!3.08\!\! \\
		SkyGP-F-1\!\! & \!0.037\! & -\!1.83  &0.26 & -\!0.92 & 0.08 & \!-\!3.29\!\! \\
		SkyGP-F-4\!\! & \!0.024\!  & -\!1.91 & 0.09 & -\!1.39 & 0.07 & \!-\!3.36\!\! \\
		SkyGP-D-1\!\! & \!0.031\! & -\!1.89  & 0.23 & -\!1.10 & 0.08& \!-\!3.31\!\! \\
		SkyGP-D-4\!\! & \!\textbf{0.017}\! & \textbf{-\!2.03}  & \textbf{0.08} & -\!1.40 &0.07& \!\textbf{-\!3.38}\!\! \\
		LocalGPs-1\!\! & \!0.031\! & -\!1.90  & 0.18 & -\!1.11 & 0.08 & \!-\!3.26\!\! \\
		LocalGPs-4\!\! &  \!0.071\! & -\!1.38  & 0.18 & {-\!0.79} & 0.14 & \!-\!2.20\!\! \\
		ISSGP\!\! & \!0.023\! & -\!1.91 & \textbf{0.08} & \textbf{-\!1.46} & \textbf{0.06} & \!-\!2.19\!\! \\
		SSGP\!\! & \!0.068\! & -\!1.11 & 0.09 & {-\!1.05} & -- & --\\
		\bottomrule
	\end{tabular}
\end{table}

\begin{table}[t]
	\centering
	\caption{Average prediction and update time on 3 dadtasets.}
	\label{tab_time}
	\begin{tabular}{lcccccc}
		\toprule
		{{Model}} 
		& \multicolumn{2}{c}{{SARCOS}} 
		& \multicolumn{2}{c}{{PUMA}} 
		& \multicolumn{2}{c}{{ELECTRIC}} \\
		\cmidrule(lr){2-3} 
		\cmidrule(lr){4-5} 
		\cmidrule(lr){6-7}
		& $t_{\text{pred}}$ & $t_{\text{up}}$ 
		& $t_{\text{pred}}$ & $t_{\text{up}}$ 
		& $t_{\text{pred}}$ & $t_{\text{up}}$ \\
		\midrule
		LoG-GP & 0.30 & 0.22  & \textbf{0.22} & 0.20 & 0.18 & 0.24 \\
		SkyGP-F-1 & \textbf{0.16} & \textbf{0.04}  &0.26 & \textbf{0.04} & 0.15 &  \textbf{0.04} \\
		SkyGP-F-4 & 0.23  &  \textbf{0.04} & 0.35 &  \textbf{0.04} & 0.21 &  \textbf{0.04} \\
		SkyGP-D-1 & 0.17 & 0.09  & 0.25 & 0.08 & \textbf{0.14}& 0.06\\
		SkyGP-D-4 & 0.24 & 0.16  & 0.28 & 0.16 &0.22& 0.08\\
		LocalGPs-1 & 1.17 & 0.15  & 1.07 & 0.08 & 2.78 & 0.06\\
		LocalGPs-4 & 1.25 & 0.23  & 1.14 & 0.08 & 2.88 & 0.06\\
		ISSGP & 18 & 7  & 3 & 6 & 4 & 8\\
		SSGP & 5 & 4 & 2 & 6 & -- & --\\
		\bottomrule
	\end{tabular}
\end{table}

%%%%%%%%%%%%%%%%%%%%%%%%%%%%%%%%%%%%%%%%%%%%%%%%%%%%%%%%%%%%%%%%
\subsection{Control Performance Evaluation}
\label{subsec_control_performance}
We evaluate the control performance of the system described as $m \ddot{q} + 9.8 = u  + f(\boldsymbol{x})$ with $m=1$, where
\begin{align}
	f(\boldsymbol{x}) =\; & 1 + x_1 x_2 / 10 + \cos(x_2) / 2 \\
	& - 10 \sin(5x_1) + (1 + \exp(-x_2/10))^{-1} / 2. \nonumber
\end{align}
Choose $x_1 = q, x_2 = \dot{q}$, then the dynamics is reformulated as \( \dot{x}_1 = x_2, \; \dot{x}_2 = g(u)  + f(\boldsymbol{x}) \) with $g(u) = u - 9.8 = (u - c \dot{q} - g) / m$ similar as \eqref{eq_ELcontroller} with $c = 0$ and $g = 9.8$.
\begin{align}
	&\pi(t, \bm{x}, \tilde{\mu}) = 9.8 - \tilde{\mu}(\bm{x}) - a_r w_r^2 \sin(w_r t) + k_p (q - q_d(t))\nonumber\\
	&\quad\quad\quad\quad~~+ k_d (\dot{q} - \dot{q}_d(t)) - [0,1] \bm{P} (\bm{x} - \bm{x}_d) / (2 \varepsilon)\!
\end{align}
with control gains $k_p = 5$, $k_d = 10$ and $\varepsilon = 1$. 
The matrix $\bm{P}$ is obtained by solving the Lyapunov equation with $\bm{Q} = \bm{I}_2$.
The two variants of SkyGP are with \( \bar{N} = 50 \), $\bar{W}=10$, \( \Gamma = 1 \) and the maximal data size set to 100. Other hyperparameters are set the same in \cref{subsec_regression_performance}.
Furthermore, the desired reference is chosen with the form $q_r(t) = a_r \sin(w_r t)$ with coefficients \( a_r=1, w_r=0.1 \). 

\begin{figure}[t]
	\centering
	\includegraphics[width=0.47\textwidth]{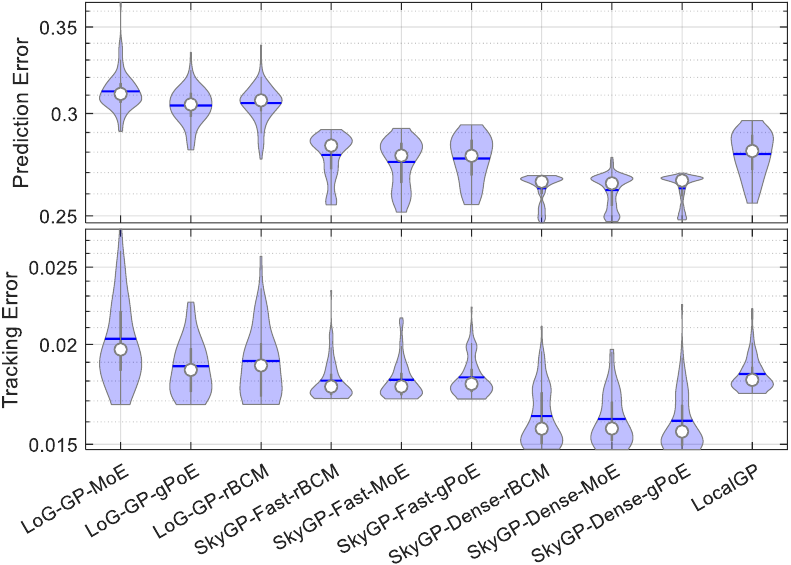}
	\caption{Learning and tracking performance comparison of the control task from the $100$ times Monte Carlo tests.}
	\label{fig_prediction_tracking}
\end{figure}

The maximum norm values of tracking errors $\bm{z} = \bm{x} - \bm{x}_d$ and prediction errors $f(\bm{x}) - \tilde{\mu}(\bm{x})$ are shown in \cref{fig_prediction_tracking}. 
The box plot summarizes the distribution of the performance on prediction error and tracking error over 100 trials with a random initial state $\bm{x}(0)$ uniformly distributed in $[0,1]^2$. 
The proposed SkyGP variants demonstrate both lower average and median in prediction and tracking errors, indicating more accurate and consistent learning and control behavior compared to all baselines. 
Particularly, the prediction and tracking errors of SkyGP-Dense are lower than those of SkyGP-Fast with sufficiently fast computation time, which is consistent with the regression benchmarks in \cref{subsec_regression_performance}. 
These results reinforce the advantages of the SkyGP framework in closed-loop dynamical systems, where both precision and safety are critical.

%%%%%%%%%%%%%%%%%%%%%%%%%%%%%%%%%%%%%%%%%%%%%%%%%%%%%%%%%%%%%%%%
%%%%%%%%%%%%%%%%%%%%%%%%%%%%%%%%%%%%%%%%%%%%%%%%%%%%%%%%%%%%%%%%
\section{Conclusion and Discussion}
\label{sec_Conclusion}
In this paper, we introduced SkyGP for scalable and adaptive learning. Designed to address the computational and memory bottlenecks of exact GPs in real-time settings, SkyGP maintains a bounded set of experts while preserving strong predictive performance with uncertainty quantification. 
SkyGP-Fast, targeted for computational efficiency, and SkyGP-Dense, designed for high prediction accuracy. 
Our extensive empirical evaluation across multiple real-world regression datasets and real-time control tasks demonstrates the superiority of SkyGP over SOTA baselines. 

A limitation of the center update rule in \cref{eq_center} is that it may fail to accurately capture special data distributions, such as annular or multimodal patterns. 
We adopt the method from~\cite{Nguyen_NeurIPS2008_Local} for its simplicity and computational efficiency. 
Another limitation lies in the use of the dynamic table list, which may struggle to handle data with sudden distributional shifts, potentially affecting the timely allocation or reuse of experts.

\section*{Acknowledgements} 
The authors acknowledge the financial support by the Federal Ministry of Education and Research of Germany in the programme of ``Souverän. Digital. Vernetzt.'' under joint project 6G-life with project identification number: 16KISK002.

%%%%%%%%%%%%%%%%%%%%%%%%%%%%%%%%%%%%%%%%%%%%%%%%%%%%%%%%%%%%%%%%
%%%%%%%%%%%%%%%%%%%%%%%%%%%%%%%%%%%%%%%%%%%%%%%%%%%%%%%%%%%%%%%%
% \clearpage
\bibliography{ref}

@article{lederer2024safe,
	title={Safe barrier-constrained control of uncertain systems via event-triggered learning},
	author={Lederer, Armin and Begzadi{\'c}, Azra and Hirche, Sandra and Cort{\'e}s, Jorge and Herbert, Sylvia},
	journal={arXiv preprint arXiv:2408.16144}       ,
	year={2024}
}

@article{hashimoto2022learning,
	title={{Learning-based Symbolic Abstractions for Nonlinear Control Systems}},
	author={Hashimoto, Kazumune and Saoud, Adnane and Kishida, Masako and Ushio, Toshimitsu and Dimarogonas, Dimos V},
	journal={{Automatica}},
	volume={146},
	pages={110646},
	year={2022},
	publisher={Elsevier}
}

@article{maddalena2021deterministic,
	title={{Deterministic Error Bounds for Kernel-based Learning Techniques under Bounded Noise}},
	author={Maddalena, Emilio Tanowe and Scharnhorst, Paul and Jones, Colin N},
	journal={{Automatica}},
	volume={134},
	pages={109896},
	year={2021},
	publisher={Elsevier}
}

@article{scharnhorst2022robust,
	title={{Robust Uncertainty Bounds in Reproducing Kernel Hilbert Spaces: A Convex Optimization Approach}},
	author={Scharnhorst, Paul and Maddalena, Emilio T and Jiang, Yuning and Jones, Colin N},
	journal={{IEEE Transactions on Automatic Control}},
	volume={68},
	number={5},
	pages={2848--2861},
	year={2022},
	publisher={IEEE}
}

@article{srinivas2012information,
	title={{Information-theoretic Regret Bounds for Gaussian Process Optimization in the Bandit Setting}},
	author={Srinivas, Niranjan and Krause, Andreas and Kakade, Sham M and Seeger, Matthias W},
	journal={{IEEE Transactions on Information Theory}},
	volume={58},
	number={5},
	pages={3250--3265},
	year={2012},
	publisher={IEEE}
}

@article{lederer2019uniform,
	title={Uniform error bounds for Gaussian process regression with application to safe control},
	author={Lederer, Armin and Umlauft, Jonas and Hirche, Sandra},
	journal={Advances in Neural Information Processing Systems},
	volume={32},
	year={2019}
}

@book{khalil2015nonlinear,
	title={{Nonlinear Systems}},
	author={Khalil, Hassan K},
	year={2002},
	publisher={{Prentice-Hall}}
}

@article{Wachi_2018AAAI_safe, 
title={{Safe Exploration and Optimization of Constrained MDPs Using Gaussian Processes}}, volume={32}, url={https://ojs.aaai.org/index.php/AAAI/article/view/12103}, DOI={10.1609/aaai.v32i1.12103}, abstractNote={ &lt;p&gt; We present a reinforcement learning approach to explore and optimize a safety-constrained Markov Decision Process(MDP). In this setting, the agent must maximize discounted cumulative reward while constraining the probability of entering unsafe states, defined using a safety function being within some tolerance. The safety values of all states are not known a priori, and we probabilistically model them via aGaussian Process (GP) prior. As such, properly behaving in such an environment requires balancing a three-way trade-off of exploring the safety function, exploring the reward function, and exploiting acquired knowledge to maximize reward. We propose a novel approach to balance this trade-off. Specifically, our approach explores unvisited states selectively; that is, it prioritizes the exploration of a state if visiting that state significantly improves the knowledge on the achievable cumulative reward. Our approach relies on a novel information gain criterion based on Gaussian Process representations of the reward and safety functions. We demonstrate the effectiveness of our approach on a range of experiments, including a simulation using the real Martian terrain data. &lt;/p&gt; }, number={1}, journal={Proceedings of the AAAI Conference on Artificial Intelligence}, author={Wachi, Akifumi and Sui, Yanan and Yue, Yisong and Ono, Masahiro}, year={2018}, month={Apr.} }

@article{yang_EAAI2025_safe,
title = {{Safe event-triggered control of unmanned surface vehicles with Gaussian processes: Resilience in denial of service attacks and uncertain dynamics}},
journal = {Engineering Applications of Artificial Intelligence},
volume = {153},
pages = {110776},
year = {2025},
issn = {0952-1976},
doi = {https://doi.org/10.1016/j.engappai.2025.110776}      ,
url = {https://www.sciencedirect.com/science/article/pii/S0952197625007766},
author = {Zewen Yang and Xiaobing Dai and Liang Fang and Jiajia Zhou and Zheping Yan}
}

@InProceedings{dai_L4DC2023_can,
  title = 	 {{Can Learning Deteriorate Control? Analyzing Computational Delays in Gaussian Process-Based Event-Triggered Online Learning}},
  author =       {Dai, Xiaobing and Lederer, Armin and Yang, Zewen and Hirche, Sandra},
  booktitle = 	 {Proceedings of The 5th Annual Learning for Dynamics and Control Conference},
  pages = 	 {445--457},
  year = 	 {2023},
  editor = 	 {Matni, Nikolai and Morari, Manfred and Pappas, George J.},
  volume = 	 {211},
  series = 	 {Proceedings of Machine Learning Research},
  month = 	 {15--16 Jun},
  publisher =    {PMLR},
  url = 	 {https://proceedings.mlr.press/v211/dai23a.html}
}

@inproceedings{Nguyen_NeurIPS2008_Local,
 author = {Nguyen-tuong, Duy and Peters, Jan and Seeger, Matthias},
 booktitle = {Advances in Neural Information Processing Systems},
 editor = {D. Koller and D. Schuurmans and Y. Bengio and L. Bottou},
 pages = {},
 publisher = {Curran Associates, Inc.},
 title = {{Local Gaussian Process Regression for Real Time Online Model Learning and Control}},
 url = {https://proceedings.neurips.cc/paper_files/paper/2008/file/01161aaa0b6d1345dd8fe4e481144d84-Paper.pdf},
 volume = {21},
 year = {2008}
}

@book{rasmussenGaussianProcessesMachine2006,
  title = {{Gaussian Processes for Machine Learning}},
  author = {Rasmussen, Carl Edward and Williams, Christopher K. I.},
  year = {2006},
  series = {Adaptive Computation and Machine Learning},
  publisher = {{MIT Press}},
  address = {{Cambridge, Mass}},
  isbn = {978-0-262-18253-9},
  lccn = {QA274.4 .R37 2006},
}

@InProceedings{lederer_ICML2021_Gaussian,
  title = 	 {{Gaussian Process-Based Real-Time Learning for Safety Critical Applications}},
  author =       {Lederer, Armin and Conejo, Alejandro J Ord{\'o}{\~n}ez and Maier, Korbinian A and Xiao, Wenxin and Umlauft, Jonas and Hirche, Sandra},
  booktitle = 	 {Proceedings of the 38th International Conference on Machine Learning},
  pages = 	 {6055--6064},
  year = 	 {2021},
  editor = 	 {Meila, Marina and Zhang, Tong},
  volume = 	 {139},
  series = 	 {Proceedings of Machine Learning Research},
  month = 	 {18--24 Jul},
  publisher =    {PMLR},
  url = 	 {https://proceedings.mlr.press/v139/lederer21a.html},
}

@article{yang_AAAI2025_Asynchronous, 
title={{Asynchronous Distributed Gaussian Process Regression}}, volume={39}, url={https://ojs.aaai.org/index.php/AAAI/article/view/34359}, DOI={10.1609/aaai.v39i21.34359}, abstractNote={In this paper, we address a practical distributed Bayesian learning problem with asynchronous measurements and predictions due to diverse computational conditions. To this end, asynchronous distributed Gaussian process (AsyncDGP) regression is proposed, which is the first effective online distributed Gaussian processes (GPs) approach to improve the prediction accuracy in real-time learning tasks. By leveraging the devised evaluation criterion and established prediction error bounds, AsyncDGP enables the distinction of contributions of each model for prediction ensembling using aggregation strategy. Furthermore, we extend its utility to dynamic systems by introducing a learning-based control law, ensuring guaranteed control performance in safety-critical applications. Additionally, a networked online learning simulation platform for distributed GPs, namely online GP gym (GPgym), is introduced for testing the performance of learning and control of dynamical systems. Numerical simulations within GPgym across regression tasks with real-world data sets and dynamical control scenarios demonstrate the effectiveness and applicability of AsyncDGP.}, number={21}, journal={Proceedings of the AAAI Conference on Artificial Intelligence}, author={Yang, Zewen and Dai, Xiaobing and Hirche, Sandra}, year={2025}, month={Apr.}, pages={22065-22073} }

@inproceedings{bui_NeurIPS2017_Streaming,
 author = {Bui, Thang D and Nguyen, Cuong and Turner, Richard E},
 booktitle = {Advances in Neural Information Processing Systems},
 editor = {I. Guyon and U. Von Luxburg and S. Bengio and H. Wallach and R. Fergus and S. Vishwanathan and R. Garnett},
 pages = {},
 publisher = {Curran Associates, Inc.},
 title = {{Streaming Sparse Gaussian Process Approximations}},
 url = {https://proceedings.neurips.cc/paper_files/paper/2017/file/f31b20466ae89669f9741e047487eb37-Paper.pdf},
 volume = {30},
 year = {2017}
}

@article{Arjan_NN2013_Real,
title = {Real-time model learning using Incremental Sparse Spectrum Gaussian Process Regression},
journal = {Neural Networks},
volume = {41},
pages = {59-69},
year = {2013},
note = {Special Issue on Autonomous Learning},
issn = {0893-6080},
doi = {https://doi.org/10.1016/j.neunet.2012.08.011}     ,
url = {https://www.sciencedirect.com/science/article/pii/S0893608012002249},
author = {Arjan Gijsberts and Giorgio Metta}
}

@inproceedings{Snelson_NeurIPS2005_Sparse,
 author = {Snelson, Edward and Ghahramani, Zoubin},
 booktitle = {Advances in Neural Information Processing Systems},
 editor = {Y. Weiss and B. Sch\"{o}lkopf and J. Platt},
 pages = {},
 publisher = {MIT Press},
 title = {{Sparse Gaussian Processes using Pseudo-inputs}},
 url = {https://proceedings.neurips.cc/paper_files/paper/2005/file/4491777b1aa8b5b32c2e8666dbe1a495-Paper.pdf},
 volume = {18},
 year = {2005}
}

@inproceedings{Naish_NeurIPS2007_Generalized,
 author = {Naish-guzman, Andrew and Holden, Sean},
 booktitle = {Advances in Neural Information Processing Systems},
 editor = {J. Platt and D. Koller and Y. Singer and S. Roweis},
 pages = {},
 publisher = {Curran Associates, Inc.},
 title = {{The Generalized FITC Approximation}},
 url = {https://proceedings.neurips.cc/paper_files/paper/2007/file/94c7bb58efc3b337800875b5d382a072-Paper.pdf},
 volume = {20},
 year = {2007}
}

@inproceedings{Reeb_NeurIPS2018_Learning,
 author = {Reeb, David and Doerr, Andreas and Gerwinn, Sebastian and Rakitsch, Barbara},
 booktitle = {Advances in Neural Information Processing Systems},
 editor = {S. Bengio and H. Wallach and H. Larochelle and K. Grauman and N. Cesa-Bianchi and R. Garnett},
 pages = {},
 publisher = {Curran Associates, Inc.},
 title = {{Learning Gaussian Processes by Minimizing PAC-Bayesian Generalization Bounds}},
 url = {https://proceedings.neurips.cc/paper_files/paper/2018/file/d43ab110ab2489d6b9b2caa394bf920f-Paper.pdf},
 volume = {31},
 year = {2018}
}

@InProceedings{Deisenroth_ICML2015_Distributed,
  title = 	 {{Distributed Gaussian Processes}},
  author = 	 {Deisenroth, Marc and Ng, Jun Wei},
  booktitle = 	 {Proceedings of the 32nd International Conference on Machine Learning},
  pages = 	 {1481--1490},
  year = 	 {2015},
  editor = 	 {Bach, Francis and Blei, David},
  volume = 	 {37},
  series = 	 {Proceedings of Machine Learning Research},
  address = 	 {Lille, France},
  month = 	 {07--09 Jul},
  publisher =    {PMLR},
  url = 	 {https://proceedings.mlr.press/v37/deisenroth15.html}
}

@article{tresp_NC2000_bayesian,
  title={{A Bayesian committee machine}},
  author={Tresp, Volker},
  journal={Neural computation},
  volume={12},
  number={11},
  pages={2719--2741},
  year={2000},
  publisher={MIT Press One Rogers Street, Cambridge, MA 02142-1209, USA journals-info~…}
}

@misc{cao2015generalizedproductexpertsautomatic,
      title={{Generalized Product of Experts for Automatic and Principled Fusion of Gaussian Process Predictions}}, 
      author={Yanshuai Cao and David J. Fleet},
      year={2015},
      eprint={1410.7827},
      archivePrefix={arXiv},
      primaryClass={cs.LG},
      url={https://arxiv.org/abs/1410.7827}, 
}

@InProceedings{Cohen_ICML2020_Healing,
  title = 	 {{Healing Products of {G}aussian Process Experts}},
  author =       {Cohen, Samuel and Mbuvha, Rendani and Marwala, Tshilidzi and Deisenroth, Marc},
  booktitle = 	 {Proceedings of the 37th International Conference on Machine Learning},
  pages = 	 {2068--2077},
  year = 	 {2020},
  editor = 	 {III, Hal Daumé and Singh, Aarti},
  volume = 	 {119},
  series = 	 {Proceedings of Machine Learning Research},
  month = 	 {13--18 Jul},
  publisher =    {PMLR},
  url = 	 {https://proceedings.mlr.press/v119/cohen20b.html}
}

@Article{Scheurch_ML2023_Correlated,
author={Sch{\"u}rch, Manuel
and Azzimonti, Dario
and Benavoli, Alessio
and Zaffalon, Marco},
title={{Correlated product of experts for sparse Gaussian process regression}},
journal={Machine Learning},
year={2023},
month={May},
day={01},
volume={112},
number={5},
pages={1411-1432},
issn={1573-0565},
doi={10.1007/s10994-022-06297-3},
url={https://doi.org/10.1007/s10994-022-06297-3} 
}

@InProceedings{liu_ICML2018_Generalized,
  title = 	 {{Generalized Robust {B}ayesian Committee Machine for Large-scale {G}aussian Process Regression}},
  author =       {Liu, Haitao and Cai, Jianfei and Wang, Yi and Ong, Yew Soon},
  booktitle = 	 {Proceedings of the 35th International Conference on Machine Learning},
  pages = 	 {3131--3140},
  year = 	 {2018},
  editor = 	 {Dy, Jennifer and Krause, Andreas},
  volume = 	 {80},
  series = 	 {Proceedings of Machine Learning Research},
  month = 	 {10--15 Jul},
  publisher =    {PMLR},
  url = 	 {https://proceedings.mlr.press/v80/liu18a.html}
}

@inproceedings{Tresp_NeurIPS2000_Mixtures,
 author = {Tresp, Volker},
 booktitle = {Advances in Neural Information Processing Systems},
 editor = {T. Leen and T. Dietterich and V. Tresp},
 pages = {},
 publisher = {MIT Press},
 title = {{Mixtures of Gaussian Processes}},
 url = {https://proceedings.neurips.cc/paper_files/paper/2000/file/9fdb62f932adf55af2c0e09e55861964-Paper.pdf},
 volume = {13},
 year = {2000}
}

@inproceedings{Yuan_NeurIPS2008_Variational,
 author = {Yuan, Chao and Neubauer, Claus},
 booktitle = {Advances in Neural Information Processing Systems},
 editor = {D. Koller and D. Schuurmans and Y. Bengio and L. Bottou},
 pages = {},
 publisher = {Curran Associates, Inc.},
 title = {{Variational Mixture of Gaussian Process Experts}},
 url = {https://proceedings.neurips.cc/paper_files/paper/2008/file/f4b9ec30ad9f68f89b29639786cb62ef-Paper.pdf},
 volume = {21},
 year = {2008}
}

@InProceedings{Trapp_AISTATS2020_Deep,
      title = 	 {{Deep Structured Mixtures of Gaussian Processes}},
  author =       {Trapp, Martin and Peharz, Robert and Pernkopf, Franz and Rasmussen, Carl Edward},
  booktitle = 	 {Proceedings of the Twenty Third International Conference on Artificial Intelligence and Statistics},
  pages = 	 {2251--2261},
  year = 	 {2020},
  editor = 	 {Chiappa, Silvia and Calandra, Roberto},
  volume = 	 {108},
  series = 	 {Proceedings of Machine Learning Research},
  month = 	 {26--28 Aug},
  publisher =    {PMLR},
  url = 	 {https://proceedings.mlr.press/v108/trapp20a.html}
}

@INPROCEEDINGS{yang_CDC2021_Distributed,
  author={Yang, Zewen and Sosnowski, Stefan and Liu, Qingchen and Jiao, Junjie and Lederer, Armin and Hirche, Sandra},
  booktitle={2021 60th IEEE Conference on Decision and Control (CDC)}, 
  title={{Distributed Learning Consensus Control for Unknown Nonlinear Multi-Agent Systems based on Gaussian Processes}}, 
  year={2021},
  volume={},
  number={},
  pages={4406-4411},
  doi={10.1109/CDC45484.2021.9683522}}

@ARTICLE{Lederer_TAC2024_cooperative,
  author={Lederer, Armin and Yang, Zewen and Jiao, Junjie and Hirche, Sandra},
  journal={IEEE Transactions on Automatic Control}, 
  title={{Cooperative Control of Uncertain Multiagent Systems via Distributed Gaussian Processes}}, 
  year={2023},
  volume={68},
  number={5},
  pages={3091-3098},
  doi={10.1109/TAC.2022.3205424}}

@INPROCEEDINGS{yang_ACC2024_cooperative,
  author={Yang, Zewen and Dong, Songbo and Lederer, Armin and Dai, Xiaobing and Chen, Siyu and Sosnowski, Stefan and Hattab, Georges and Hirche, Sandra},
  booktitle={2024 American Control Conference (ACC)}, 
  title={{Cooperative Learning with Gaussian Processes for Euler-Lagrange Systems Tracking Control Under Switching Topologies}}, 
  year={2024},
  volume={},
  number={},
  pages={560-567},
  doi={10.23919/ACC60939.2024.10644832}}

@inproceedings{yang_AAMAS2024_whom,
author = {Yang, Zewen and Dai, Xiaobing and Dubey, Akshat and Hirche, Sandra and Hattab, Georges},
title = {{Whom to Trust? Elective Learning for Distributed Gaussian Process Regression}},
year = {2024},
isbn = {9798400704864},
publisher = {International Foundation for Autonomous Agents and Multiagent Systems},
address = {Richland, SC},
booktitle = {Proceedings of the 23rd International Conference on Autonomous Agents and Multiagent Systems},
pages = {2020–2028},
numpages = {9},
location = {Auckland, New Zealand},
series = {AAMAS '24}
}

@article{dai_EJC2024_Decentralized,
title = {{Decentralized event-triggered online learning for safe consensus control of multi-agent systems with Gaussian process regression}},
journal = {European Journal of Control},
volume = {80},
pages = {101058},
year = {2024},
note = {2024 European Control Conference Special Issue},
issn = {0947-3580},
doi = {https://doi.org/10.1016/j.ejcon.2024.101058},
url = {https://www.sciencedirect.com/science/article/pii/S0947358024001183},
author = {Xiaobing Dai and Zewen Yang and Mengtian Xu and Sihua Zhang and Fangzhou Liu and Georges Hattab and Sandra Hirche}
}

@ARTICLE{dai_TNNLS2025_Cooperative,
  author={Dai, Xiaobing and Yang, Zewen and Zhang, Sihua and Zhai, Di-Hua and Xia, Yuanqing and Hirche, Sandra},
  journal={IEEE Transactions on Neural Networks and Learning Systems}, 
  title={{Cooperative Online Learning for Multiagent System Control via Gaussian Processes With Event-Triggered Mechanism}}, 
  year={2025},
  volume={36},
  number={7},
  pages={13304-13318},
  doi={10.1109/TNNLS.2024.3446732}}

@ARTICLE{Yuan_TAC2024_Lightweight,
  author={Yuan, Zhenyuan and Zhu, Minghui},
  journal={IEEE Transactions on Automatic Control}, 
  title={{Lightweight Distributed Gaussian Process Regression for Online Machine Learning}}, 
  year={2024},
  volume={69},
  number={6},
  pages={3928-3943},
  doi={10.1109/TAC.2024.3351555}}

\clearpage
\appendix

%%%%%%%%%%%%%%%%%%%%%%%%%%%%%%%%%%%%%%%%%%%%%%%%%%%%%%%%%%%%%%%%
%%%%%%%%%%%%%%%%%%%%%%%%%%%%%%%%%%%%%%%%%%%%%%%%%%%%%%%%%%%%%%%%
\section{Proofs}
\label{sec_proofs}
\subsection{Proof of \cref{lemma:gp-error-bound}}
    \begin{proof}
		Following the prediction error bound for single GP expert in \cite{lederer2019uniform}, the error between $f(\cdot)$ and $\mu_i(\cdot)$ is probabilistic bounded by
		\begin{align} \label{eqn_single_GP_error_bound}
			| f(\bm{x}) - \mu_i(\bm{x}) | \leq \beta \sigma_i(\bm{x}) + \gamma_i, && \forall \bm{x} \in \mathbb{X}
		\end{align}
		with a probability of at least $1 - \delta$.
		Considering that the weights $\omega_i(\cdot)$ satisfy $\sum_{i \in \mathcal{N}} \omega_i(\bm{x}) = 1$, the aggregated prediction error is written as
		\begin{align}
			| f(\bm{x}) - \tilde{\mu}_i(\bm{x}) | =& | f(\bm{x}) - \sum\nolimits_{i \in \mathcal{N}} \omega_i(\bm{x}) \mu_i(\bm{x}) | \\
			\le& \sum\nolimits_{i \in \mathcal{N}} \omega_i(\bm{x}) | f(\bm{x}) - \mu_i(\bm{x}) |.
		\end{align}
		Apply the prediction error bound in \eqref{eqn_single_GP_error_bound}
		\begin{align}
			| f(\bm{x}) - \tilde{\mu}_i(\bm{x}) | \le& \sum\nolimits_{i \in \mathcal{N}} \omega_i(\bm{x}) (\beta \sigma_i(\bm{x}) + \gamma_i) \\
			=& \beta \sum\nolimits_{i \in \mathcal{N}} \omega_i(\bm{x}) \sigma_i(\bm{x}) + \sum\nolimits_{i \in \mathcal{N}} \omega_i(\bm{x}) \gamma_i \nonumber
		\end{align}
		with a probability of at least $1 - | \mathcal{N} | \delta$ using the union bound.
		Considering $| \mathcal{N} | \le \bar{\mathcal{N}}$, the result in \cref{lemma:gp-error-bound} is derived, concluding the proof.
	\end{proof}
    
\subsection{Proof of \cref{theorem_boundZ}}
\begin{proof}
	Note that the difference between $\bm{f}(\cdot, \cdot)$ and $\bm{\mu}(\cdot, \cdot)$ leads to the bounded time derivative of $V(\cdot, \cdot)$ as
	\begin{align}
		\dot{V}(t, \bm{z}) \!\!=& \nabla_t V(t, \bm{x}) \\
		&+ \nabla_{\bm{x}} V(t, \bm{x}) (\bm{f}(\bm{x}, \bm{\pi}(t, \bm{x}, \bm{\mu})) + \bm{g}(\bm{x}, \bm{\pi}(t, \bm{x}, \bm{\mu}))) \nonumber \\
		\le& \nabla_t V(t, \bm{x}) \\
		&+ \nabla_{\bm{x}} V(t, \bm{x}) (\bm{\mu}(\bm{x}, \bm{\pi}(t, \bm{x}, \bm{\mu})) + \bm{g}(\bm{x}, \bm{\pi}(t, \bm{x}, \bm{\mu}))) \nonumber \\
		&+\! \| \nabla_{\bm{x}} V(t,\! \bm{x}) \| \| \bm{f}(\bm{x},\! \bm{\pi}(t,\! \bm{x},\! \bm{\mu})) \!\!-\!\! \bm{\mu}(\bm{x},\! \bm{\pi}(t,\! \bm{x},\! \bm{\mu})) \| \nonumber \\
		\le& - \alpha( V(t, \bm{z}) ) - \| \nabla_{\bm{x}} V(t, \bm{x}) \|^2 / (4 \varepsilon) \\
		&+ \| \nabla_{\bm{x}} V(t, \bm{x}) \| \| \bm{\eta}(\bm{\xi}(t, \bm{x}, \bm{\mu})) \| \nonumber
	\end{align}
	where $\bm{\xi}(t, \bm{x}, \bm{\mu}) = [\bm{x}^T, \bm{\pi}^T(t, \bm{x}, \bm{\mu})]^T$ and $\bm{\eta}(\cdot) = [\eta_1(\cdot), \cdots, \eta_{d_x}(\cdot)]^T$ with $\eta_i(\cdot)$ obtained using prediction error bound with kernel $\kappa_i(\cdot, \cdot)$.
	Moreover, applying Young's inequality with $\varepsilon \in \mathbb{R}_+$ on the third term inducing
	\begin{align}
		\| \nabla_{\bm{x}} V(t, \bm{x}) \| \| \bm{\eta}(\bm{\xi}(t, \bm{x}, \bm{\mu})) \| \le& \| \nabla_{\bm{x}} V(t, \bm{x}) \|^2 / (4 \varepsilon) \\
		&+ \varepsilon \| \bm{\eta}(\bm{\xi}(t, \bm{x}, \bm{\mu})) \|^2, \nonumber
	\end{align}
	then the bound of $\dot{V}(t, \bm{z})$ is further written as
	\begin{align}
		\dot{V}(t, \bm{z}) \le& - \alpha( V(t, \bm{z}) ) + \varepsilon \| \bm{\eta}(\bm{\xi}(t, \bm{x}, \bm{\mu})) \|^2 \\
		\le& - \alpha( V(t, \bm{z}) ) + \varepsilon \bar{\eta}^2
	\end{align}
	with $\bar{\eta} = \max_{t \in \mathbb{R}_{0,+}, \bm{x} \in \mathbb{X}} \| \bm{\eta}(\bm{\xi}(t, \bm{x}, \bm{\mu})) \|$.
	Therefore, it is easy to derive the ultimate bound of $V(t, \bm{z}(t))$ as
	\begin{align}
		\lim_{t \to \infty} V(t, \bm{z}(t)) \le \alpha^{-1}(\varepsilon \bar{\eta}^2),
	\end{align}
	which leads to the bound of $\bm{z}(t)$ as
	\begin{align}
		\lim_{t \to \infty} \| \bm{z}(t) \| \le& \lim_{t \to \infty} \underline{\alpha}^{-1}(V(t, \bm{z}(t))) \\
		\le& \underline{\alpha}^{-1}(\alpha^{-1}(\varepsilon \bar{\eta}^2)),
	\end{align}
	concluding the proof.
\end{proof}

\subsection{Proof of \cref{theorem_tracking}}
\begin{proof}
It is easy to see the existence of class-$\mathcal{K}$ functions $\underline{\alpha}(\cdot)$ and $\bar{\alpha}(\cdot)$ as
\begin{align}
	\underline{\alpha}(\| \bm{z} \|) = \underline{\lambda}(\bm{P}) \| \bm{z} \|^2, &&
	\bar{\alpha}(\| \bm{z} \|) = \bar{\lambda}(\bm{P}) \| \bm{z} \|^2
\end{align}
using the control law \eqref{eq_ELcontroller}.
Moreover, apply \eqref{eq_ELcontroller} to \eqref{eq_ELsystem}, the controlled dynamics is written as
\begin{align}
	\ddot{\bm{q}} &= \bm{M}^{-1}(\bm{q}) ( \bm{u} - \bm{C}(\bm{x}) \dot{\bm{q}} - \bm{g}(\bm{x}) + \bm{d}(\bm{x}) ) \nonumber\\
	&= \bm{M}^{-1}(\bm{q}) ( \bm{d}(\bm{x}) - \bm{\mu}(\bm{x}) - \bm{B}^T(\bm{x}) \bm{P} \bm{z} / (2 \varepsilon)) \nonumber\\
	&\quad + [\bm{K}_p, \bm{K}_d] \bm{z} + \ddot{\bm{q}}_d(t),
\end{align}
such that the dynamics w.r.t $\bm{z}$ becomes
\begin{align}
	\dot{\bm{z}} 
	= \bm{A} \bm{z} - \bm{B}(\bm{x}) \bm{B}^T(\bm{x}) \bm{P} \bm{z} / (2 \varepsilon) + \bm{B}(\bm{x}) ( \bm{d}(\bm{x}) - \bm{\mu}(\bm{x}) ). \nonumber
\end{align}
Investigate the time derivative of $V(t, \bm{z})$ as
	\begin{align}
		\dot{V}(t, \bm{z}) =& \bm{z}^T (\bm{A}^T \bm{P} + \bm{P} \bm{A}) \bm{z} + 2 \bm{z}^T \bm{P} \bm{B}(\bm{x}) ( \bm{d}(\bm{x}) - \bm{\mu}(\bm{x}) ) \nonumber \\
		&- 2 \bm{z}^T \bm{P} \bm{B}(\bm{x}) \bm{B}^T(\bm{x}) \bm{P} \bm{z} / (2 \varepsilon) \\
		\le& - \bm{z}^T \bm{Q} \bm{z} + \| \bm{B}^T(\bm{x}) \bm{P} \bm{z} \|^2 / \varepsilon + \varepsilon \| \bm{d}(\bm{x}) - \bm{\mu}(\bm{x}) \|^2 \nonumber \\
		&- \| \bm{B}^T(\bm{x}) \bm{P} \bm{z} \|^2 / \varepsilon
	\end{align}
	using Young's inequality.
	Moreover, considering the relaxation with eigenvalues and prediction error bound for GP, the upper bound of $\dot{V}(t, \bm{z})$ becomes
	\begin{align}
		\dot{V}(t, \bm{z}) \le& - \underline{\lambda}(\bm{Q}) \| \bm{z} \|^2 + \varepsilon \| \bm{\eta}(\bm{x}) \|^2 \\
		\le& - \underline{\lambda}(\bm{Q}) \bar{\lambda}^{-1}(\bm{P}) V(t, \bm{z}) + \varepsilon \bar{\eta}^2.
	\end{align}
	with a probability of at least $1 - d_q \delta$.
	According to the previous result, it is directly derived that
	\begin{align}
		\lim_{t \to \infty} \| \bm{z}(t) \| \le \frac{ \varepsilon \bar{\lambda}(\bm{P}) }{\underline{\lambda}(\bm{P}) \underline{\lambda}(\bm{Q})}  \bar{\eta}^2,
	\end{align}
	which concludes the proof.
\end{proof}

%%%%%%%%%%%%%%%%%%%%%%%%%%%%%%%%%%%%%%%%%%%%%%%%%%%%%%%%%%%%%%%%
%%%%%%%%%%%%%%%%%%%%%%%%%%%%%%%%%%%%%%%%%%%%%%%%%%%%%%%%%%%%%%%%
\section{Additional Results}
\label{sec_additional_results}

\subsection{Simulation Setup}
The codes are executed in Python 3.10.12 on the laptop with Intel i5-12500H  with 16.0GB RAM.

\subsection{Regression Benchmark}
We evaluated the performance of the LoG-GP, LocalGPs, ISSGP, SSGP, using the following parameters:
\begin{itemize}
    \item LoG-GP: Each expert has $K = 2$ children with $\bar{\mathcal{N}}=50$.
    \item LocalGP: Each expert is with $\bar{\mathcal{N}}=50$. We use the divided threshold as 0.9 in~\cite{Nguyen_NeurIPS2008_Local}.
    \item ISSGP: We use $D = 200$ random Fourier features with as the same in~\cite{Arjan_NN2013_Real}.
    \item SSGP: We use 100 inducing points for initial optimization. For online optimization of hyperparameters and inducing points, we use  the position of the inducing points every 500 samples using 50 optimization steps per batch, in order to balance both predictive accuracy and computational efficiency.
\end{itemize}

We use the average prediction and update times as a performance metric, as well as the standardized mean squared error (SMSE) and the mean standardized log loss (MSLL) following the sequential evaluation protocol described in~\cite{rasmussenGaussianProcessesMachine2006}.
\begin{equation}
\text{SMSE}_k = \frac{1}{k s_y^2} \sum_{i=1}^k \left( \hat{\mu}_{k-1}(\boldsymbol{x}^{i}) - y^{i} \right)^2,
\label{eq:smse}
\end{equation}
where \( \hat{\mu}_{k-1}(\boldsymbol{x}^{i}) \) denotes the predictive mean after observing the first \( k{-}1 \) data points, and \( s_y^2 \) is the empirical variance of the ground truth labels \( y^{i} \).

In \cref{tab_average_time_4}, it is shown that the proposed SkyGP-Fast variant, consistently achieves the lowest model update times of 0.04 ms on both the SARCOS and ELECTRIC datasets and the lowest prediction times of 0.16 ms and 0.14 ms, respectively. 
Although LoGGP attains the fastest prediction time on the KIN40K and PUMA datasets, it falls short in model update time compared to SkyGP-Fast. 
Notably, all SkyGP variants maintain better performance across all datasets, even as the number of active experts increases, compared to the baselines, i.e., SSGP, ISSGP, and LocalGPs, which exhibit significantly higher computational overhead. 
The SkyGP-Dense variant provides improvements in predictive accuracy due to its data replacement strategy, while incurring only a marginal increase in computational cost—making it a compelling option when accuracy is prioritized. 
Overall, the results highlight SkyGP’s superior scalability and practical utility for streaming online learning.

\begin{table*}[t]
\centering
\caption{Average prediction and update time (ms) across datasets.}
\label{tab_average_time_4}
\begin{tabular}{lcccccccc}
\toprule
{\textbf{Model}} 
& \multicolumn{2}{c}{\textbf{SARCOS}} 
& \multicolumn{2}{c}{\textbf{KIN40K}} 
& \multicolumn{2}{c}{\textbf{PUMA}} 
& \multicolumn{2}{c}{\textbf{ELECTRIC}} \\
\cmidrule(lr){2-3} 
\cmidrule(lr){4-5} 
\cmidrule(lr){6-7} 
\cmidrule(lr){8-9}
& $t_{\text{pred}}$ & $t_{\text{up}}$ 
& $t_{\text{pred}}$ & $t_{\text{up}}$ 
& $t_{\text{pred}}$ & $t_{\text{up}}$ 
& $t_{\text{pred}}$ & $t_{\text{up}}$ \\
\midrule
LoGGP-MOE & 0.28 & {0.21} & \textbf{0.24} & 0.16 & 0.23 & 0.18 & 0.19 & 0.25 \\
LoGGP-gPOE & 0.29 & 0.20 & \textbf{0.24} & 0.16 & \textbf{0.22} & 0.19 & 0.22 & 0.24 \\
LoGGP-rBCM & 0.30 & 0.22 & \textbf{0.24} & 0.16 & \textbf{0.22} & 0.20 & 0.18 & 0.24 \\
SkyGP-Fast-MoE ($\bar{\mathcal{N}}=1$) & 0.17 & \textbf{0.04} & 0.30 & \textbf{0.04} & 0.29 & 0.04 & 0.15 & \textbf{0.04} \\
SkyGP-Fast-MoE ($\bar{\mathcal{N}}=2$) & 0.20 & \textbf{0.04} & 0.32 & \textbf{0.04} & 0.33 & 0.04 & 0.17 & \textbf{0.04}\\
SkyGP-Fast-MoE ($\bar{\mathcal{N}}=4$) & 0.23 & \textbf{0.04} & 0.35 & \textbf{0.04} & 0.34 & 0.04 & 0.20 & \textbf{0.04}\\
SkyGP-Dense-MoE ($\bar{\mathcal{N}}=1$) & 0.18 & 0.09 & 0.38 & 0.06 & 0.30 & 0.08 &\textbf{0.14}& 0.06\\
SkyGP-Dense-MoE ($\bar{\mathcal{N}}=2$) & 0.19 & 0.12 & 0.29 & 0.09 & 0.30 & 0.12 &0.15& 0.07\\
SkyGP-Dense-MoE ($\bar{\mathcal{N}}=4$) & 0.23 & 0.16 & 0.32 & 0.11 & 0.32 & 0.16 &0.19& 0.08\\
SkyGP-Fast-gPoE ($\bar{\mathcal{N}}=1$) & \textbf{0.16} & \textbf{0.04} & 0.30 & \textbf{0.04} & 0.29 & \textbf{0.04} & 0.15 & \textbf{0.04} \\
SkyGP-Fast-gPoE ($\bar{\mathcal{N}}=2$) & 0.19 & \textbf{0.04} & 0.35 & \textbf{0.04} & 0.33 & \textbf{0.04} & 0.17 & \textbf{0.04}\\
SkyGP-Fast-gPoE ($\bar{\mathcal{N}}=4$) & 0.23 & \textbf{0.04} & 0.36 & \textbf{0.04} & 0.35 & \textbf{0.04} & 0.20 &\textbf{ 0.04}\\
SkyGP-Dense-gPoE ($\bar{\mathcal{N}}=1$) & 0.18 & 0.09 & 0.28 & 0.06 & 0.31 & 0.08 &\textbf{0.14}& 0.06\\
SkyGP-Dense-gPoE ($\bar{\mathcal{N}}=2$) & 0.21 & 0.12 & 0.29 & 0.09 & 0.29 & 0.12 &0.17& 0.07\\
SkyGP-Dense-gPoE ($\bar{\mathcal{N}}=4$) & 0.24 & 0.16 & 0.35 & 0.11 & 0.32 & 0.16 &0.21& 0.08\\
SkyGP-Fast-rBCM ($\bar{\mathcal{N}}=1$) & \textbf{0.16} & \textbf{0.04} & 0.30 & \textbf{0.04} & 0.26 & \textbf{0.04} & 0.15 & \textbf{0.04} \\
SkyGP-Fast-rBCM ($\bar{\mathcal{N}}=2$) & 0.18 & \textbf{0.04} & 0.33 & \textbf{0.04} & 0.28 & \textbf{0.04} & 0.19 & \textbf{0.04} \\
SkyGP-Fast-rBCM ($\bar{\mathcal{N}}=4$) & 0.23 & \textbf{0.04} & 0.35 & \textbf{0.04} & 0.31 & \textbf{0.04} & 0.21 & \textbf{0.04} \\
SkyGP-Dense-rBCM ($\bar{\mathcal{N}}=1$) & 0.17 & 0.09 & 0.28 & 0.06 & 0.25 & 0.08 &\textbf{0.14}& 0.06\\
SkyGP-Dense-rBCM ($\bar{\mathcal{N}}=2$) & 0.18 & 0.12 & 0.30 & 0.09 & 0.25 & 0.12 &0.16& 0.07\\
SkyGP-Dense-rBCM ($\bar{\mathcal{N}}=4$) & 0.24 & 0.16 & 0.35 & 0.11 & 0.28 & 0.16 &0.22& 0.08\\
LocalGPs ($\bar{\mathcal{N}}=1$) & 1.17 & 0.15 & 2.32 & 0.05 & 1.07 & 0.08 & 2.78 & 0.06\\
LocalGPs ($\bar{\mathcal{N}}=2$) & 1.27 & 0.17 & 2.07 & 0.05 & 1.07 & 0.08 & 2.67 & 0.06\\
LocalGPs ($\bar{\mathcal{N}}=4$) & 1.25 & 0.23 & 2.03 & 0.05 & 1.14 & 0.08 & 2.88 & 0.06\\
ISSGP & 18 & 7 & 3 & 8 & 3 & 6 & 4 & 8\\
SSGP & 5 & 4 & 3 & 8 & 2 & 6 & - & -\\
\bottomrule
\end{tabular}
\end{table*}

\begin{figure*}
    \centering
    \includegraphics[width=1\linewidth]{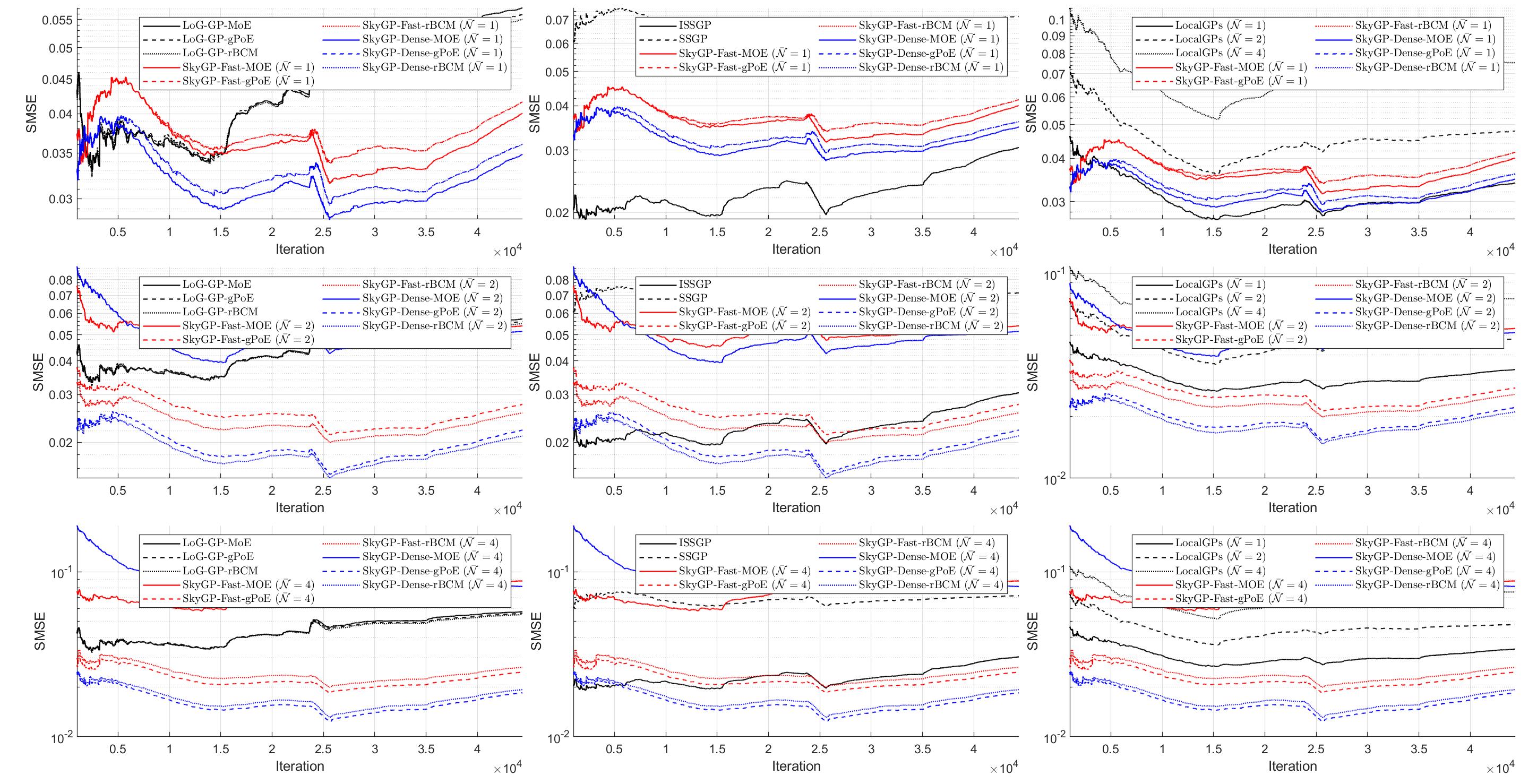}
    \caption{SMSE comparison on SARCOS dataset against all baselines.}
    \label{fig_sarcos_smse}
\end{figure*}

\begin{figure*}
    \centering
    \includegraphics[width=1\linewidth]{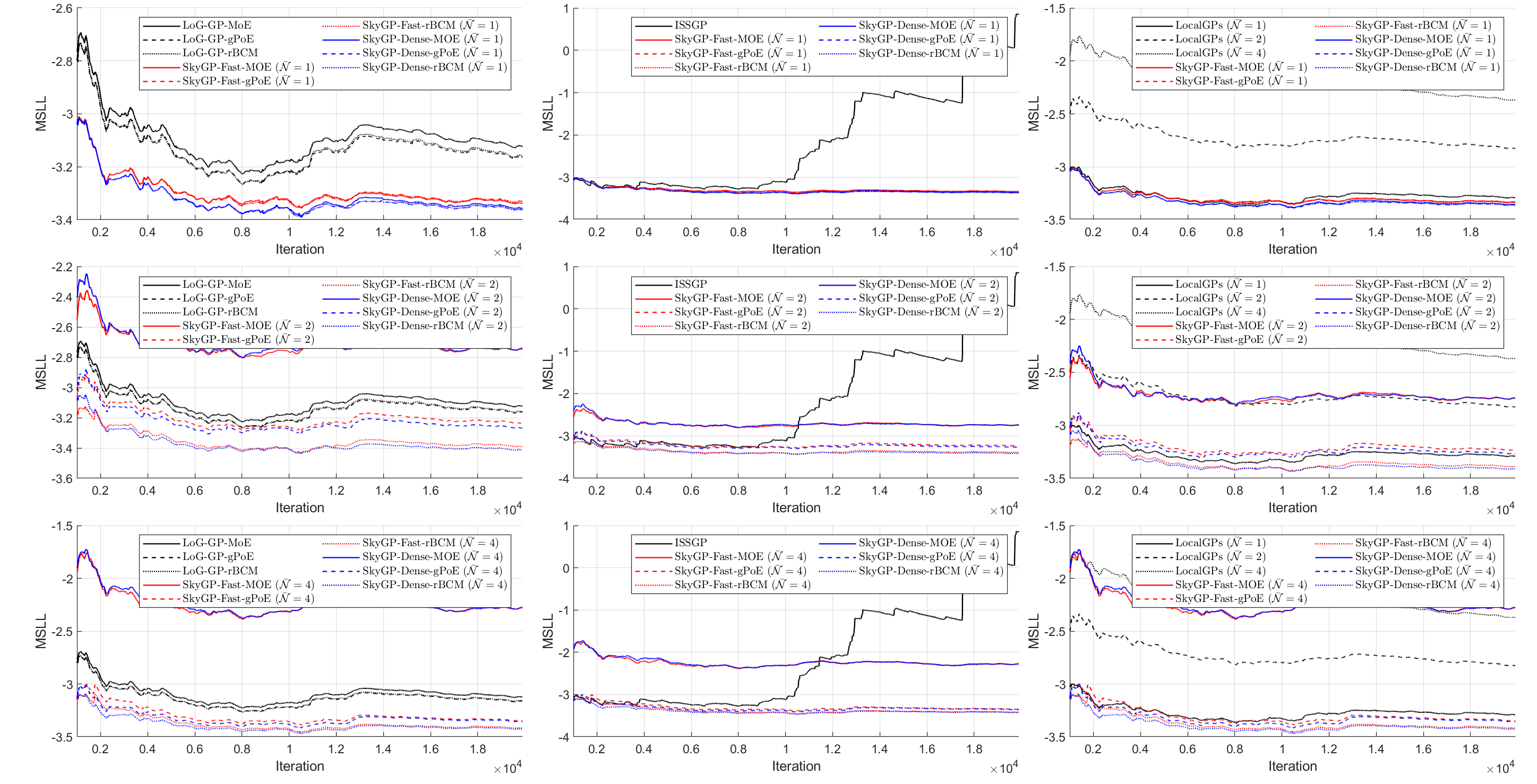}
    \caption{MSLL comparison on ELECTRIC dataset against all baselines.}
    \label{fig_puma_msll}
\end{figure*}

To visualize the performance comparison across all methods, we plot the SMSE and MSLL trends over simulation iterations in \cref{fig_sarcos_smse} to \ref{fig_puma_msll}. 
On the SARCOS dataset, both SkyGP-Fast and SkyGP-Dense with $\bar{\mathcal{N}} = 2, 4$ consistently achieve lower SMSE than the state-of-the-art LoGGP method. 
While SkyGP-Fast with $\bar{\mathcal{N}} = 1$ initially exhibits higher error before iteration 1500, it surpasses LoGGP in accuracy in later stages. 
Similarly, SkyGP-Dense outperforms LoGGP after around 500 iterations. Compared to SSGP, only the SkyGP-Dense variants with $\bar{\mathcal{N}} = 2, 4$ achieve better performance, however, SSGP suffers from high computational costs that hinder its practical use in real-time scenarios. 
ISSGP performs the worst among all evaluated methods on the SARCOS dataset. 
While LocalGPs with $\bar{\mathcal{N}} = 1$ show comparable or slightly better performance than SkyGP-Fast with the same number of experts, both SkyGP-Fast and SkyGP-Dense using gPoE and rBCM aggregation strategies outperform LocalGPs when $\bar{\mathcal{N}} = 2, 4$.

Similar to the observations on the SARCOS dataset, both SkyGP-Fast and SkyGP-Dense with $\bar{\mathcal{N}} = 2, 4$ consistently achieve lower SMSE on the PUMA dataset compared to the state-of-the-art LoGGP and LocalGPs methods. 
While SSGP and ISSGP outperform all SkyGP variants with $\bar{\mathcal{N}} = 1$, SkyGP-Dense with $\bar{\mathcal{N}} = 4$ surpasses ISSGP and achieves comparable performance to SSGP. 
As with LoGGP, LocalGPs generally perform better than SkyGP when $\bar{\mathcal{N}} = 1$. 
However, although LocalGPs with $\bar{\mathcal{N}} = 2$ show slightly better or comparable results to SkyGP-Fast with the same number of experts, both SkyGP-Fast and SkyGP-Dense outperform LocalGPs when $\bar{\mathcal{N}} = 4$.
\begin{figure*}
    \centering 
    \includegraphics[width=1\linewidth]{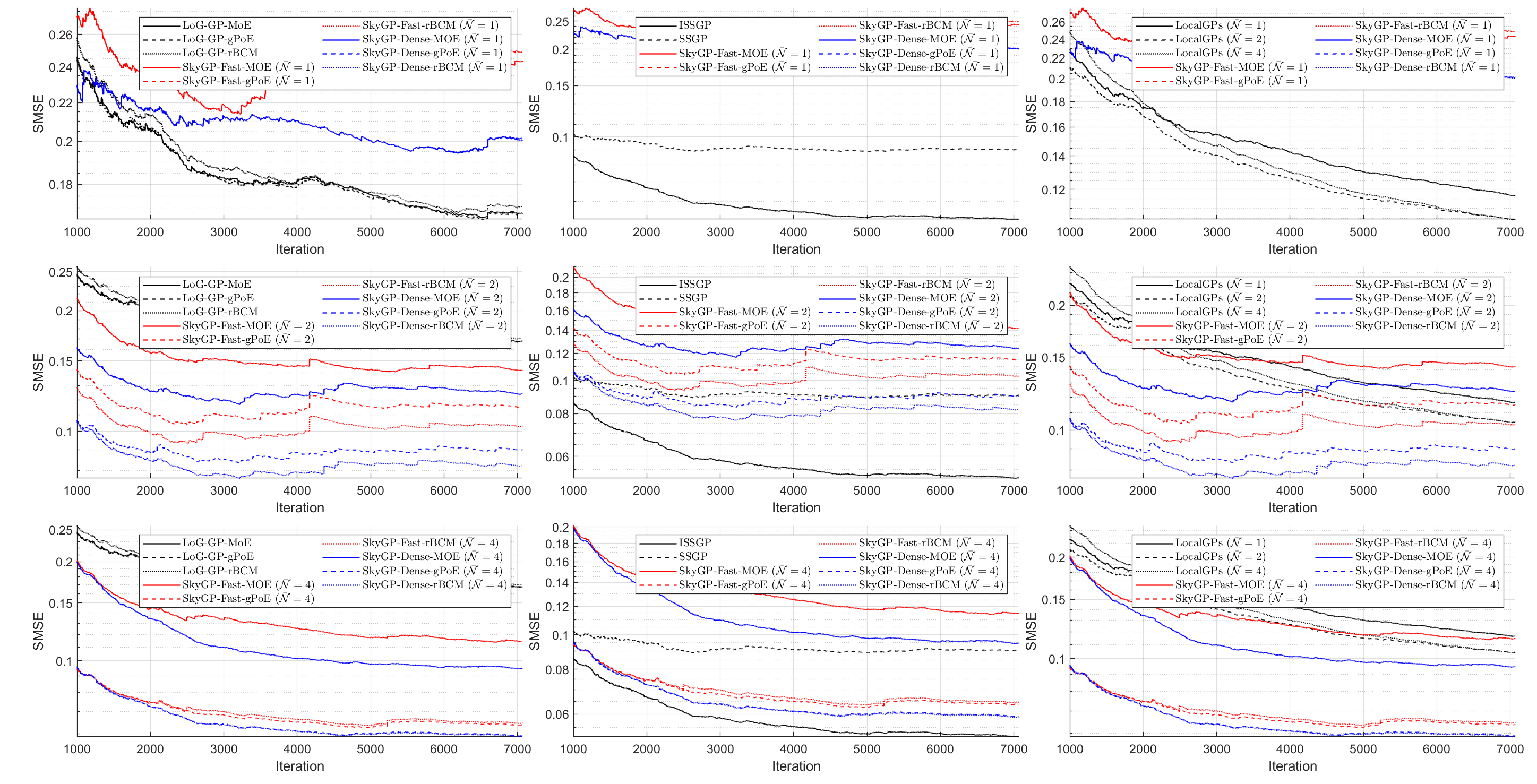}
    \caption{SMSE comparison on PUMA dataset against all baselines.}
\end{figure*}

In the KIN40K dataset, SkyGP with a single expert does not outperform the baseline methods. However, when $\bar{\mathcal{N}} = 4$, SkyGP-Dense with gPoE and rBCM aggregation strategies achieves superior performance compared to all other methods. SkyGP-Fast with $\bar{\mathcal{N}} = 4$ outperforms SSGP and shows comparable results to ISSGP.
\begin{figure*}
    \centering
    \includegraphics[width=1\linewidth]{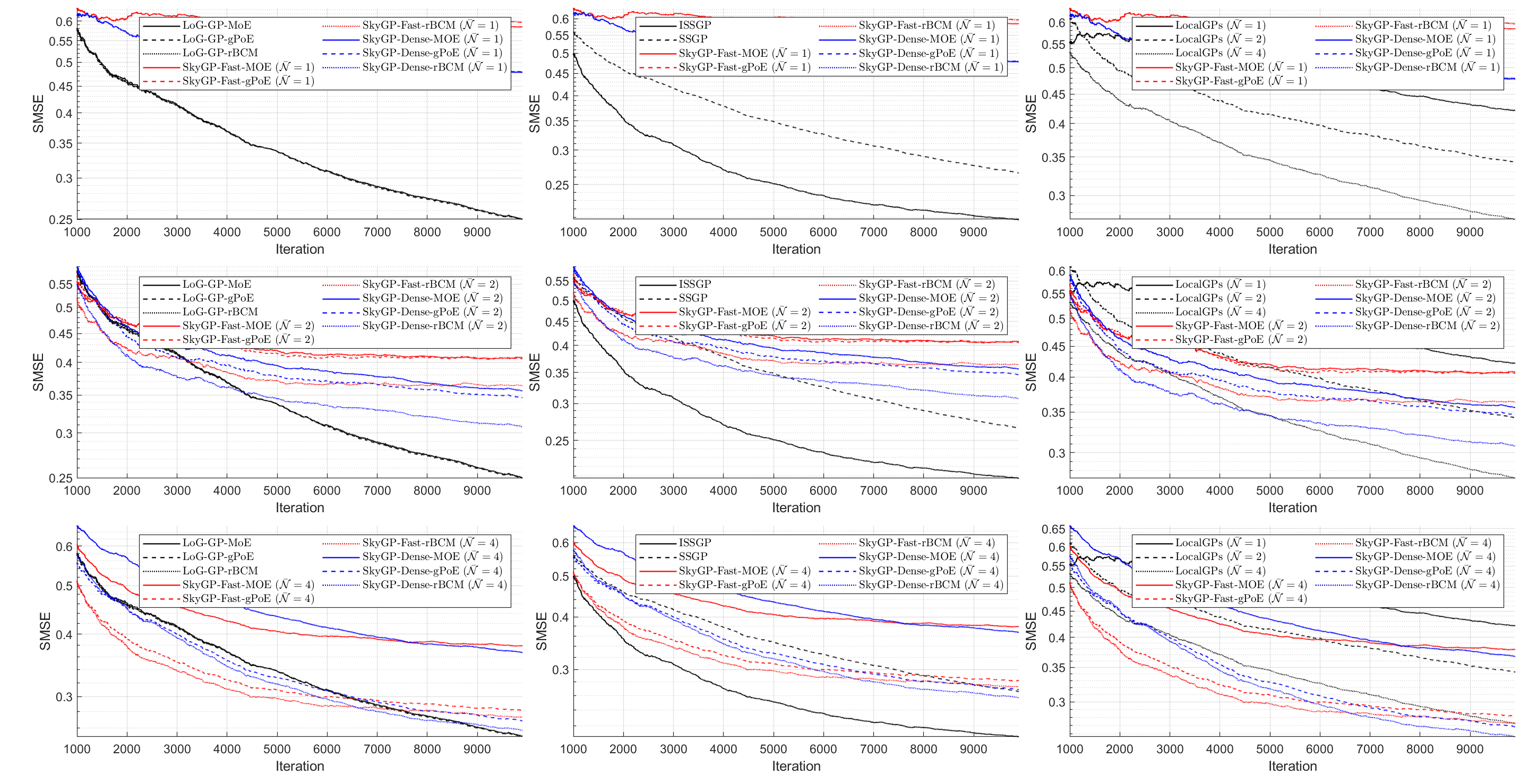}
    \caption{SMSE comparison on KIN40K dataset against all baselines.}
\end{figure*}

On the ELECTRIC dataset, SkyGP with a single expert ($\bar{\mathcal{N}} = 1$) outperforms LoG-GP but falls short compared to ISSGP, and performs similarly to LocalGPs with the same number of experts. However, when the number of experts increases to $\bar{\mathcal{N}} = 2$ or $4$, only the gPoE and rBCM aggregation strategies in SkyGP lead to improved performance over LoG-GP. A similar pattern is observed in LocalGPs, where only certain aggregation strategies with $\bar{\mathcal{N}} = 2, 4$ yield competitive or superior results compared to LoG-GP. 
Notably, due to the excessively high prediction and model update time of SSGP on the ELECTRIC dataset, we omit its results from the simulation.
\begin{figure*}
    \centering
    \includegraphics[width=1\linewidth]{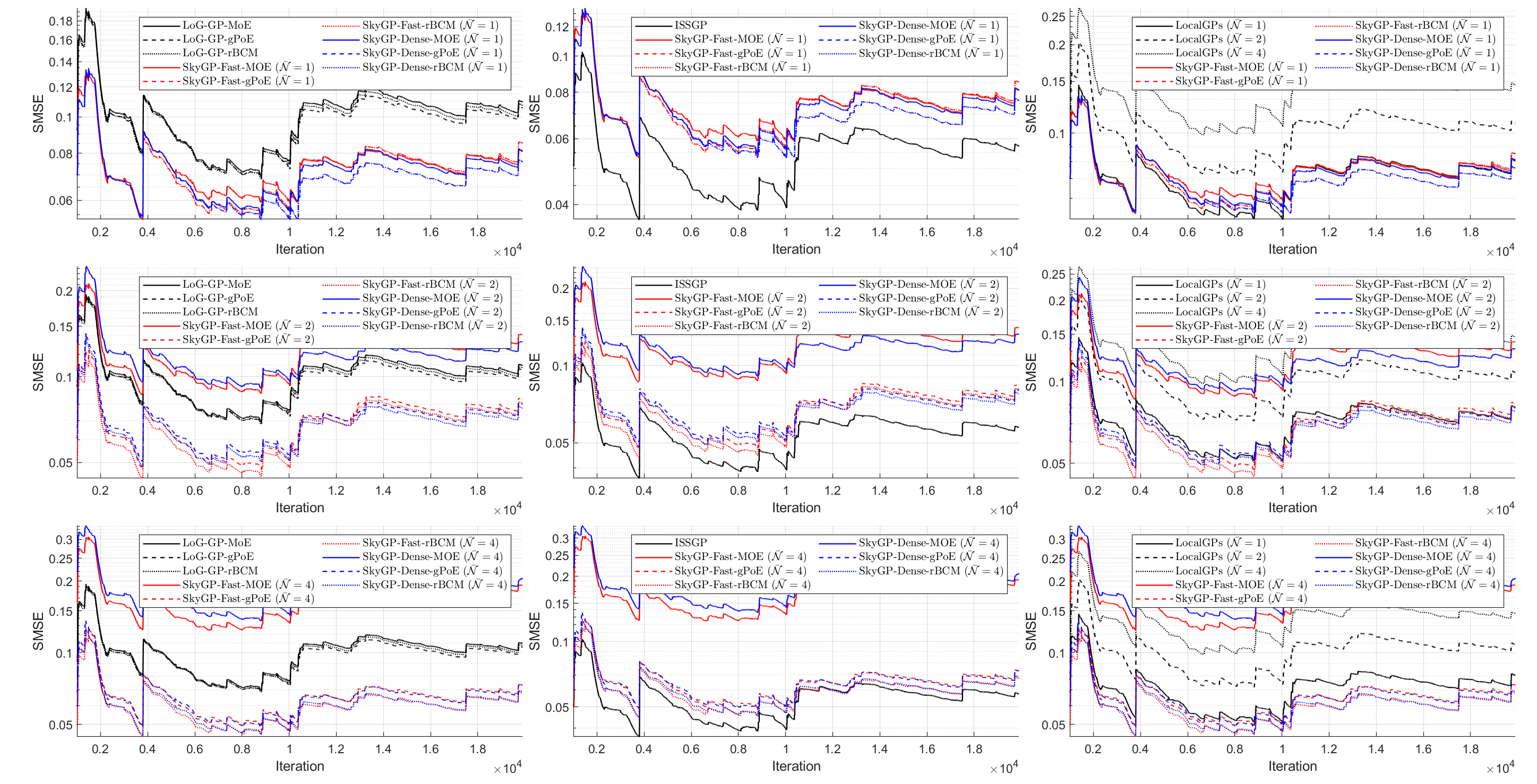}
    \caption{SMSE comparison on ELECTRIC dataset against all baselines.}
\end{figure*}

In summary, SkyGP variants demonstrate consistently strong prediction performance across multiple datasets, particularly when using multiple experts ($\bar{\mathcal{N}} = 2, 4$) and advanced aggregation strategies such as gPoE and rBCM. 
On the SARCOS and PUMA datasets, both SkyGP-Fast and SkyGP-Dense outperform the SOTA LoGGP and LocalGPs methods in terms of SMSE, especially as the number of experts increases. 
Although SkyGP with a single expert underperforms compared to SSGP and ISSGP, SkyGP-Dense with $\bar{\mathcal{N}} = 4$ achieves comparable or superior results while maintaining computational efficiency. 
On the KIN40K dataset, SkyGP-Dense with four experts and appropriate aggregation outperforms all baselines, while SkyGP-Fast matches ISSGP and surpasses SSGP. 
In the ELECTRIC dataset, SkyGP with one expert performs better than LoGGP but lags behind ISSGP. Only gPoE and rBCM aggregation in both SkyGP and LocalGPs lead to meaningful improvements when the number of experts increases. 
Overall, SkyGP achieves a favorable balance between accuracy and scalability, particularly with multiple experts and efficient aggregation.

\begin{figure*}
    \centering
    \includegraphics[width=1\linewidth]{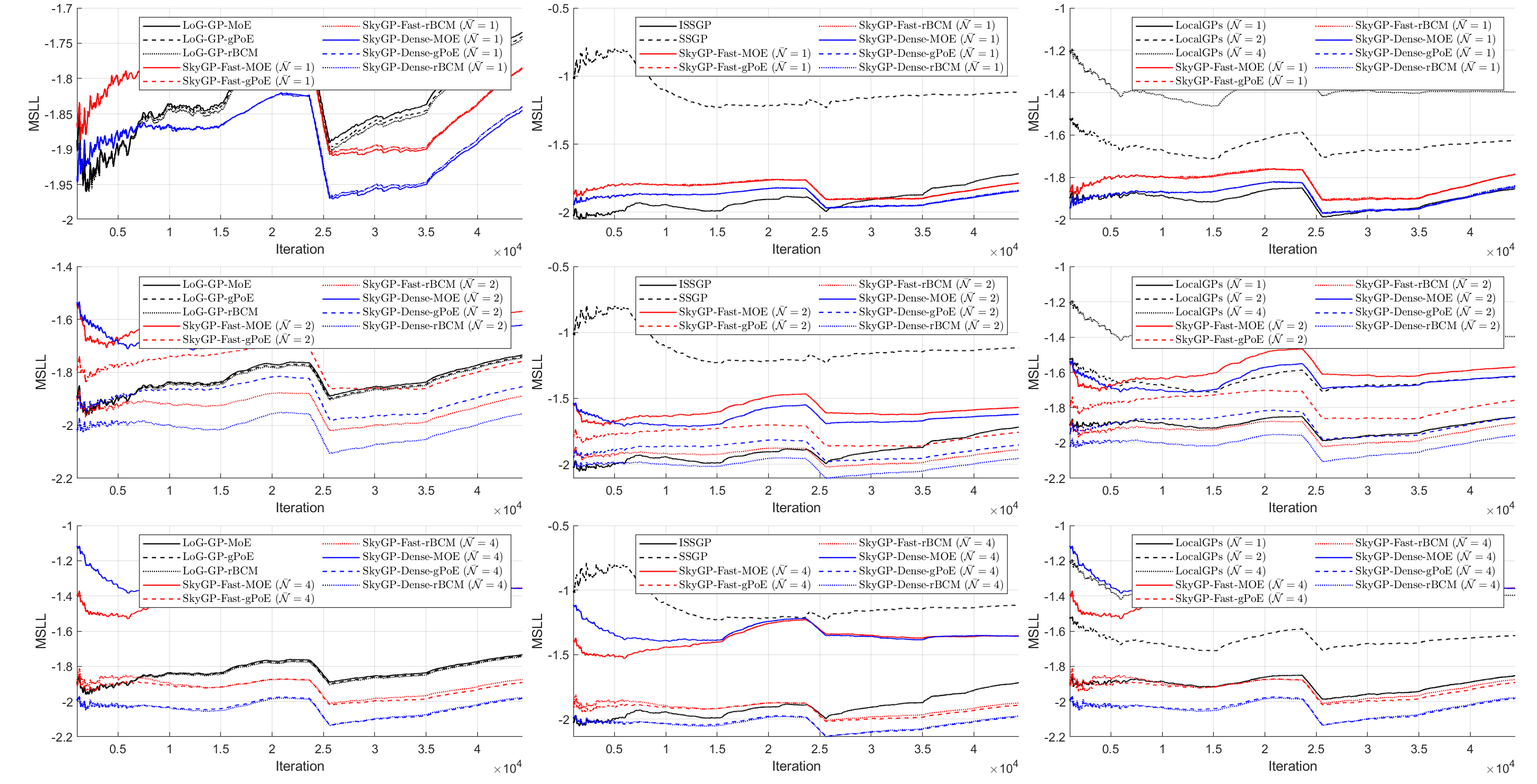}
    \caption{MSLL comparison on SARCOS dataset against all baselines.}
\end{figure*}
\begin{figure*}
    \centering
    \includegraphics[width=1\linewidth]{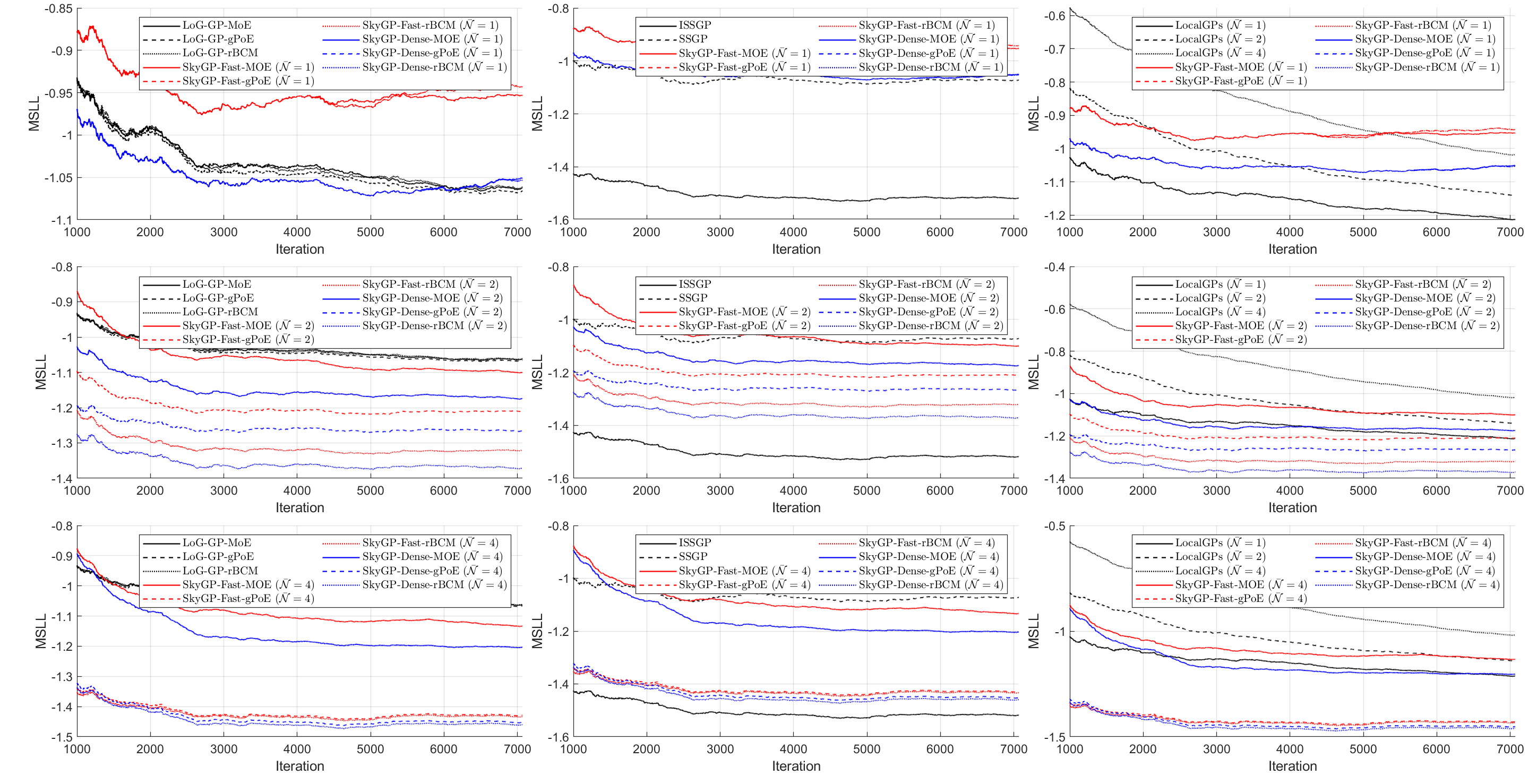}
    \caption{MSLL comparison on PUMA dataset against all baselines.}
\end{figure*}
\begin{figure*}
    \centering
    \includegraphics[width=1\linewidth]{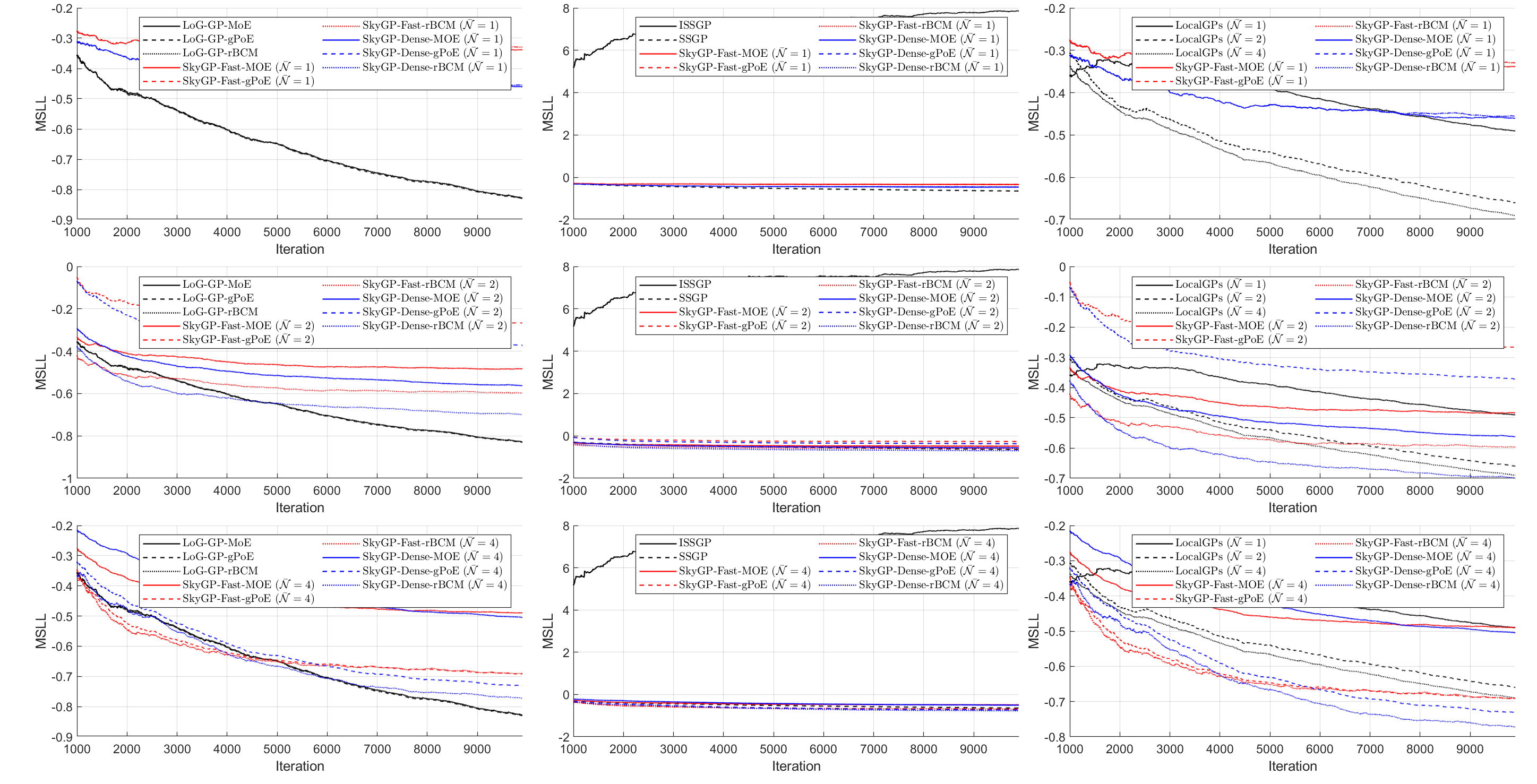}
    \caption{MSLL comparison on KIN40K dataset against all baselines.}
\end{figure*}

In terms of MSLL on the SARCOS dataset, SkyGP-Dense with $\bar{\mathcal{N}} = 4$ outperforms all baseline models, including SkyGP-Fast. SkyGP-Fast surpasses LoG-GP after approximately 500 iterations and outperforms ISSGP after around 2500 iterations, showing comparable performance to LocalGPs. 
When $\bar{\mathcal{N}} = 2$, SkyGP-Dense with the rBCM aggregation strategy achieves lower MSLL than LoG-GP, SSGP, and LocalGPs, and begins to outperform ISSGP after 2500 iterations. Notably, while both SkyGP-Dense and SkyGP-Fast outperform LoG-GP after 500 and 2500 iterations, respectively. 
Only SkyGP-Dense demonstrates competitive performance relative to LocalGPs. 
On the PUMA dataset, SkyGP variants outperform LoG-GP when using $\bar{\mathcal{N}} = 2$ or $4$, and even SkyGP-Dense with a single expert ($\bar{\mathcal{N}} = 1$) achieves better performance than LoG-GP. Compared to SSGP, all SkyGP variants demonstrate similar performance, though they do not surpass ISSGP. When $\bar{\mathcal{N}} = 2$ or $4$, both SkyGP-Dense and SkyGP-Fast outperform LocalGPs, whereas with only one expert, they perform worse than LocalGPs.
On the KIN40K dataset, all SkyGP variants outperform the sparse GP approaches (SSGP and ISSGP). 
Compared to LoG-GP, SkyGP-Fast with gPoE and rBCM achieves better performance before 2500 iterations, and SkyGP-Dense outperforms it before 4500 iterations when using 2 experts. 
With $\bar{\mathcal{N}} = 4$, SkyGP-Fast consistently outperforms LoG-GP before 5000 iterations. 
When compared to LocalGPs, SkyGP with $\bar{\mathcal{N}} = 4$ demonstrates superior performance, and SkyGP-Dense with 2 experts also outperforms LocalGPs, while SkyGP-Fast with 2 experts only shows an advantage before 5000 iterations.
On the ELECTRIC dataset, when $\bar{\mathcal{N}} = 1$, all variants of SkyGP outperform the baseline models, including LoG-GP, SSGP, ISSGP, and LocalGPs. When using 2 or 4 experts with gPoE or rBCM aggregation, SkyGP continues to outperform LoG-GP, ISSGP, and SSGP. 
However, when compared to LocalGPs, only SkyGP-Fast and SkyGP-Dense with rBCM achieve better performance when $\bar{\mathcal{N}} = 2$.

Across all datasets, SkyGP-Dense consistently achieves the best MSLL performance, particularly with $\bar{\mathcal{N}} = 4$, outperforming both SkyGP-Fast and all baseline models. 
SkyGP-Fast also shows better results with 2 or 4 experts, often surpassing LoG-GP, SSGP, and ISSGP. 
Notably, on ELECTRIC and PUMA, even single-expert SkyGP-Dense ($\bar{\mathcal{N}} = 1$) performs better than LoG-GP. 
And on KIN40K and ELECTRIC, all SkyGP variants outperform the sparse baselines.

%%%%%%%%%%%%%%%%%%%%%%%%%%%%%%%%%%%%%%%%%%%%%%%%%%%%%%%%%%%%%%%%
\subsection{Control Tasks}
\begin{figure}[t]
    \centering
    \includegraphics[width=1\linewidth]{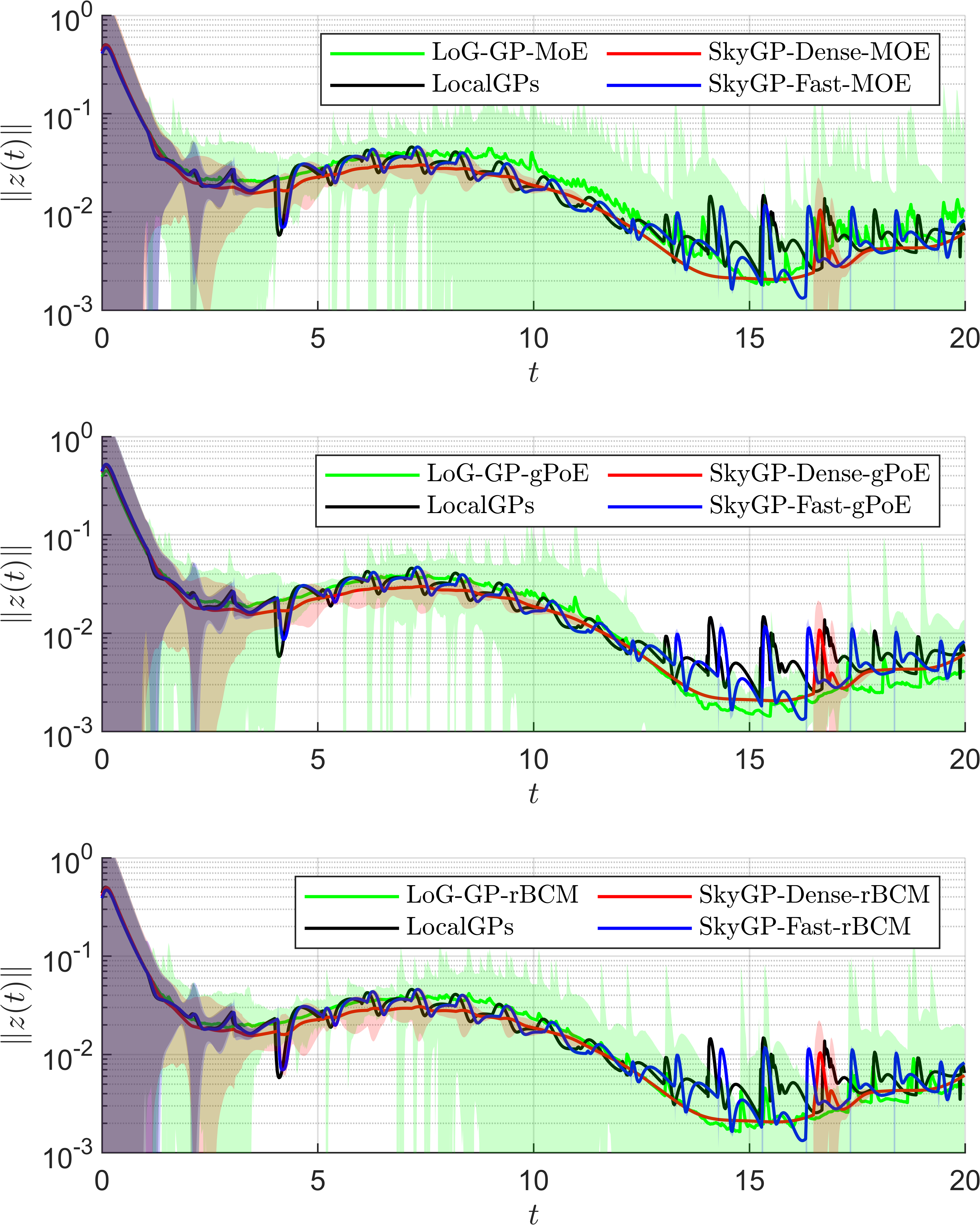}
    \caption{Tracking error comparison against LoG-GP and LocalGPs.}
    \label{fig_control_tracking_error}
\end{figure}

\begin{figure}[t]
    \centering
    \includegraphics[width=1\linewidth]{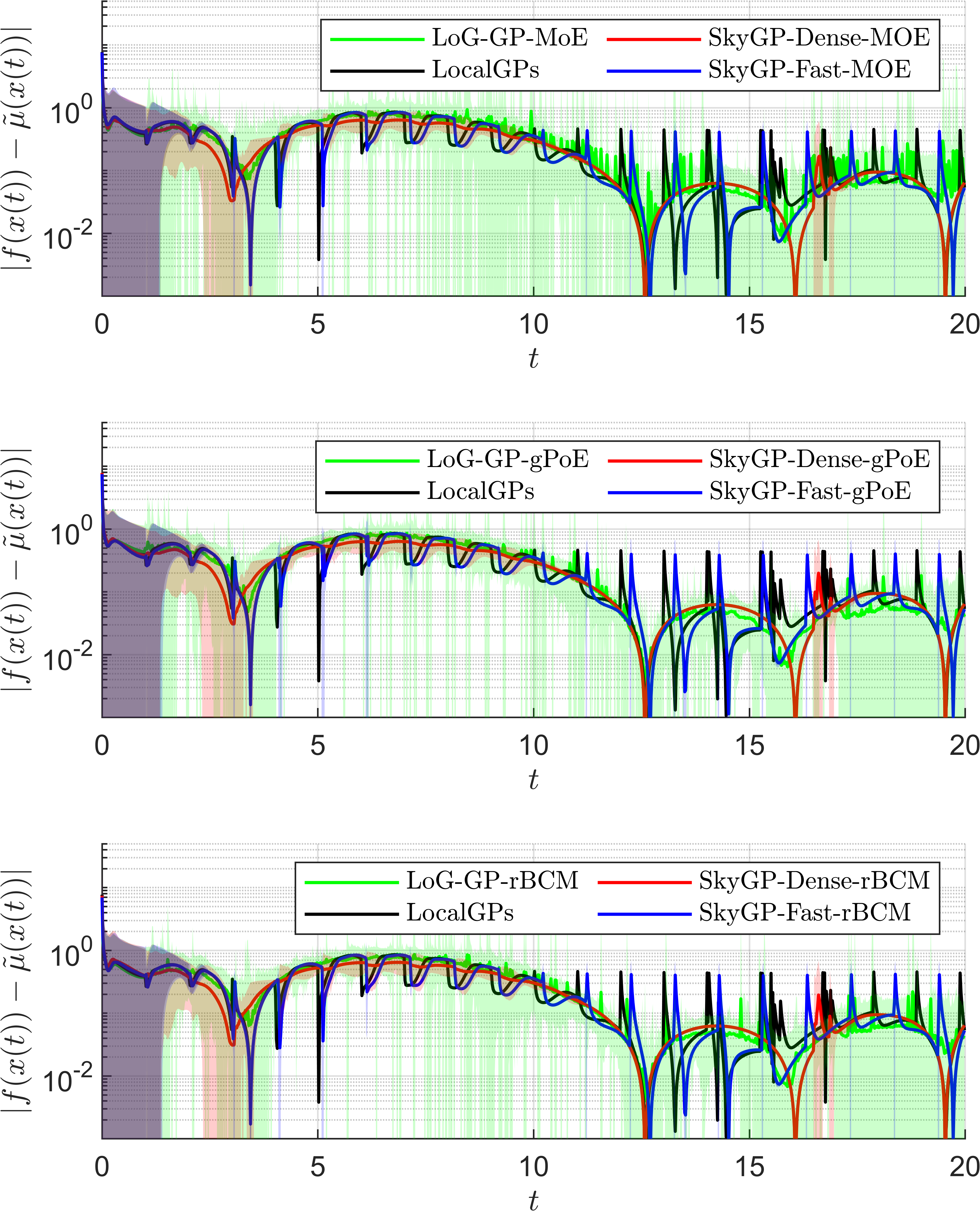}
    \caption{Prediction error comparison against LoG-GP and LocalGPs.}
    \label{fig_control_prediction_error}
\end{figure}

In the control task, we compare the baseline model with LoG-GP and LocalGP, as sparse GP methods are too slow to complete prediction and model updates within the simulation time step of 0.02 s. 
Moreover, both LoG-GP and SkyGP variants employ four experts for the aggregation method.
As shown in \cref{fig_control_tracking_error}, LoG-GP shows the largest variance and tracking error throughout the simulation.  
SkyGP-Dense achieves the lowest tracking error after approximately 13 s and demonstrates smoother tracking behavior. Moreover, SkyGP-Fast delivers performance comparable to that of LocalGPs. 
In \cref{fig_control_prediction_error}, SkyGP-Fast shows prediction errors similar to those of LocalGPs. Similarly, SkyGP-Dense maintains a smooth error profile but displays increased variance between 2 s and 3 s. In contrast, LoG-GP performs the worst in terms of both mean and variance of the prediction error.


%%%%%%%%%%%%%%%%%%%%%%%%%%%%%%%%%%%%%%%%%%%%%%%%%%%%%%%%%%%%%%%%
\subsection{Ablation Experiments}
\label{subsec_Ablation}
In this subsection, we perform ablation studies to evaluate the contribution of various algorithmic components to the overall performance of the proposed framework. 
In particular, we examine the effects of employing Euclidean distance (ED) as the expert center denoted by
\begin{align}
    \boldsymbol{c}_i^k = (k-1) \boldsymbol{c}_i^{k-1}/k + \boldsymbol{x}^k 
\end{align}
versus kernel distance (KD) in \eqref{eq_center}; 
the use of drop center (DC) ${^{\text{off}}\boldsymbol{c}_{\text{nr}}}$, 
the adaptive window mechanism (AW) in the line 2 of \cref{alg:local-expert-search-insert}, 
and the time-aware decay (TD) weighting factor $\rho$ in \cref{alg_predict_update}. 

For ED versus KD, our experimental results consistently demonstrate that KD similarity achieves superior performance in both SMSE and MSLL metrics, highlighting its effectiveness in enhancing predictive accuracy and uncertainty quantification in online settings. 
To evaluate the effect of DC, we compare models that incorporate dedicated DCs with those that rely solely on expert center distances for data replacement decisions. 
On the SARCOS and PUMA datasets, using DC generally yields better performance than not using it, while on the ELECTRIC and KIN40K datasets, the difference is negligible. 
Overall, the results indicate that incorporating drop centers can enhance predictive accuracy without adding computational overhead.
Regarding the AW strategy, we observe that disabling it yields slightly better accuracy across all datasets but increases inference latency by approximately 0.1 ms per sample. Adjusting the AW scaling factor thus enables a flexible trade-off between prediction quality and efficiency.
For the TD mechanism, results on SARCOS show consistent improvement in both SMSE and MSLL, while PUMA sees gains in SMSE but only marginal changes in MSLL. For KIN40K and ELECTRIC, TD has minimal impact on either metric.

\begin{table*}[t]
\centering
\caption{Average SMSE on the SARCOS dataset. \cmarkgreen indicates that the strategy is used; \xmarkred indicates that it is not.}
\label{tab_sarcos_smse_ablation}
\begin{tabular}{lcccccccc}
\toprule
{Strategy} 
& \multicolumn{2}{c}{{KD}} 
& \multicolumn{2}{c}{{DC}} 
& \multicolumn{2}{c}{{AW}} 
& \multicolumn{2}{c}{{TD}} \\
\cmidrule(lr){2-3} 
\cmidrule(lr){4-5} 
\cmidrule(lr){6-7} 
\cmidrule(lr){8-9}
& \cmarkgreen & \xmarkred 
& \cmarkgreen & \xmarkred  
& \cmarkgreen & \xmarkred  
& \cmarkgreen & \xmarkred  \\
\midrule
SkyGP-Fast-MoE ($\bar{\mathcal{N}}=1$) & \textbf{0.039} & 0.057 & - & - & 0.038 & \textbf{0.033} & \textbf{0.038} & \textbf{0.038} \\
SkyGP-Fast-MoE ($\bar{\mathcal{N}}=2$) & \textbf{0.057} & 0.077 & - & - & \textbf{0.050} & \textbf{0.050} & \textbf{0.056} & 0.057\\
SkyGP-Fast-MoE ($\bar{\mathcal{N}}=4$) & \textbf{0.078} & 0.097 & - & - & \textbf{0.078} & 0.085 & \textbf{0.078} & 0.084\\
SkyGP-Dense-MoE ($\bar{\mathcal{N}}=1$) & \textbf{0.033} & 0.045 & \textbf{0.033} & \textbf{0.033} & \textbf{0.033} & \textbf{0.033} & \textbf{0.033}& \textbf{0.033}\\
SkyGP-Dense-MoE ($\bar{\mathcal{N}}=2$) & \textbf{0.052} & 0.058 & \textbf{0.053} & 0.059 & 0.052 & \textbf{0.050} &\textbf{0.052}& 0.054\\
SkyGP-Dense-MoE ($\bar{\mathcal{N}}=4$) & \textbf{0.100} & 0.100 & 0.100 & \textbf{0.097} & 0.100 & \textbf{0.085} &0.099& \textbf{0.098}\\
SkyGP-Fast-gPoE($\bar{\mathcal{N}}=1$) & \textbf{0.039} & 0.057 & - & - & 0.038 & \textbf{0.033} & \textbf{0.038} & 0.038 \\
SkyGP-Fast-gPoE ($\bar{\mathcal{N}}=2$) & \textbf{0.026} & 0.033 & - & - & 0.025 & \textbf{0.020} & \textbf{0.025} & 0.026\\
SkyGP-Fast-gPoE ($\bar{\mathcal{N}}=4$) & \textbf{0.022} & 0.026 & - & - & 0.022 & \textbf{0.017} & \textbf{0.022} & 0.023\\
SkyGP-Dense-gPoE ($\bar{\mathcal{N}}=1$) & \textbf{0.033} & 0.045 & \textbf{0.030} & \textbf{0.030} & \textbf{0.033} & \textbf{0.033} & \textbf{0.033} & \textbf{0.033}\\
SkyGP-Dense-gPoE ($\bar{\mathcal{N}}=2$) & \textbf{0.019} & 0.022 & \textbf{0.019} & 0.022 & \textbf{0.019} & \textbf{0.019} &\textbf{0.020}& \textbf{0.020}\\
SkyGP-Dense-gPoE ($\bar{\mathcal{N}}=4$) & \textbf{0.017} & 0.018 & \textbf{0.017} & 0.019 & 0.017 & \textbf{0.016} &\textbf{0.017}& \textbf{0.017}\\
SkyGP-Fast-rBCM ($\bar{\mathcal{N}}=1$) & \textbf{0.039} & 0.057 & - & - & 0.038 & \textbf{0.033} & \textbf{0.038} & \textbf{0.038} \\
SkyGP-Fast-rBCM ($\bar{\mathcal{N}}=2$) & \textbf{0.023} & 0.031 & - & - & 0.023 & \textbf{0.019} & \textbf{0.023} & 0.024\\
SkyGP-Fast-rBCM ($\bar{\mathcal{N}}=4$) & \textbf{0.024} & 0.026 & - & - & 0.024 & \textbf{0.018} & \textbf{0.024} & 0.025\\
SkyGP-Dense-rBCM($\bar{\mathcal{N}}=1$) & \textbf{0.033} & 0.045 & \textbf{0.030} & \textbf{0.030} & \textbf{0.033} & \textbf{0.033} &\textbf{0.033}& \textbf{0.033}\\
SkyGP-Dense-rBCM ($\bar{\mathcal{N}}=2$) & \textbf{0.018} & 0.021 & \textbf{0.019} & 0.022 & \textbf{0.018} & \textbf{0.018} &\textbf{0.018}& 0.019\\
SkyGP-Dense-rBCM ($\bar{\mathcal{N}}=4$) & \textbf{0.018} & \textbf{0.018} & \textbf{0.018} & 0.019 & \textbf{0.017} & \textbf{0.017} &\textbf{0.017}& 0.018\\
\bottomrule
\end{tabular}
\end{table*}

\begin{table*}[t]
\centering
\caption{Average MSLL on the SARCOS dataset. \cmarkgreen indicates that the strategy is used; \xmarkred indicates that it is not.}
\label{tab_sarcos_MSLL_ablation}
\begin{tabular}{lcccccccc}
\toprule
{Strategy} 
& \multicolumn{2}{c}{{KD}} 
& \multicolumn{2}{c}{{DC}} 
& \multicolumn{2}{c}{{AW}} 
& \multicolumn{2}{c}{{TD}} \\
\cmidrule(lr){2-3} 
\cmidrule(lr){4-5} 
\cmidrule(lr){6-7} 
\cmidrule(lr){8-9}
& \cmarkgreen & \xmarkred 
& \cmarkgreen & \xmarkred  
& \cmarkgreen & \xmarkred  
& \cmarkgreen & \xmarkred  \\
\midrule
SkyGP-Fast-MoE ($\bar{\mathcal{N}}=1$) & \textbf{-1.82} & -1.74 & - & - & {-1.82} & \textbf{-1.89} & \textbf{-1.82} & \textbf{-1.82} \\
SkyGP-Fast-MoE ($\bar{\mathcal{N}}=2$) & \textbf{-1.54} & -1.42 & - & - & -1.54 & \textbf{-1.66} & \textbf{-1.54} & \textbf{-1.54}\\
SkyGP-Fast-MoE ($\bar{\mathcal{N}}=4$) & \textbf{-1.32} & -1.22 & - & - & -1.32 & \textbf{-1.38} & \textbf{-1.32} & -1.31\\
SkyGP-Dense-MoE ($\bar{\mathcal{N}}=1$) & \textbf{-1.88} & -1.82 & \textbf{-1.88} & \textbf{-1.88} & {-1.88} & \textbf{-1.89} & -1.88& \textbf{-1.89}\\
SkyGP-Dense-MoE ($\bar{\mathcal{N}}=2$) & \textbf{-1.60} & -1.57 & \textbf{-1.60} & -1.56 & -1.59 & \textbf{-1.65} &\textbf{-1.59}& \textbf{-1.59}\\
SkyGP-Dense-MoE ($\bar{\mathcal{N}}=4$) & \textbf{-1.24} & -1.23 & -1.24 & \textbf{-1.26} & -1.24 & \textbf{-1.38} &\textbf{-1.25}& \textbf{-1.25}\\
SkyGP-Fast-PoE($\bar{\mathcal{N}}=1$) & \textbf{-1.82} & -1.74 & - & - & -1.82 & \textbf{-1.89} & \textbf{-1.82} & \textbf{-1.82} \\
SkyGP-Fast-gPoE ($\bar{\mathcal{N}}=2$) & \textbf{-1.79} & -1.74 & - & - & -1.79 & \textbf{-1.88} & \textbf{-1.79} & \textbf{-1.79}\\
SkyGP-Fast-gPoE ($\bar{\mathcal{N}}=4$) & \textbf{-1.93} & -1.89 & - & - & -1.93 & \textbf{-2.04} & \textbf{-1.93} & -1.92\\
SkyGP-Dense-gPoE ($\bar{\mathcal{N}}=1$) & \textbf{-1.88} & -1.81 & \textbf{-1.88} & \textbf{-1.88} & -1.88 & \textbf{-1.89} &-1.88& \textbf{-1.89}\\
SkyGP-Dense-gPoE ($\bar{\mathcal{N}}=2$) & \textbf{-1.89} & -1.85 & \textbf{-1.89} & -1.85 & \textbf{-1.89} & -1.88 &\textbf{-1.89}& \textbf{-1.89}\\
SkyGP-Dense-gPoE ($\bar{\mathcal{N}}=4$) & \textbf{-2.01} & -2.00 & \textbf{-2.01} & -1.99 & -2.01 & \textbf{-2.03} &\textbf{-2.01}& \textbf{-2.01}\\
SkyGP-Fast-rBCM ($\bar{\mathcal{N}}=1$) & \textbf{-1.82} & -1.74 & - & - & -1.83 & \textbf{-1.89} & \textbf{-1.83} & -1.82 \\
SkyGP-Fast-rBCM ($\bar{\mathcal{N}}=2$) & \textbf{-1.93} & -1.87 & - & - & -1.93 & \textbf{-2.02} & \textbf{-1.93} & -1.92\\
SkyGP-Fast-rBCM ($\bar{\mathcal{N}}=4$) & \textbf{-1.91} & -1.89 & - & - & -1.91 & \textbf{-2.04} & \textbf{-1.91} & -1.90\\
SkyGP-Dense-rBCM($\bar{\mathcal{N}}=1$) & \textbf{-1.88} & -1.81 & \textbf{-1.88} & \textbf{-1.88} & {-1.88} & \textbf{-1.89} &\textbf{-1.89}& \textbf{-1.89}\\
SkyGP-Dense-rBCM ($\bar{\mathcal{N}}=2$) & \textbf{-2.00} & -1.98 & \textbf{-2.00} & -1.97 & -1.89 & \textbf{-2.01} &\textbf{-2.00}& -1.99\\
SkyGP-Dense-rBCM ($\bar{\mathcal{N}}=4$) & \textbf{-2.01} & -2.00 & \textbf{-2.01} & -1.99 & -2.01 & \textbf{-2.03} &\textbf{-2.01}& \textbf{-2.01}\\
\bottomrule
\end{tabular}
\end{table*}

\begin{table*}[t]
\centering
\caption{Average SMSE on the PUMA dataset. \cmarkgreen indicates that the strategy is used; \xmarkred indicates that it is not.}
\begin{tabular}{lcccccccc}
\toprule
{Strategy} 
& \multicolumn{2}{c}{{KD}} 
& \multicolumn{2}{c}{{DC}} 
& \multicolumn{2}{c}{{AW}} 
& \multicolumn{2}{c}{{TD}} \\
\cmidrule(lr){2-3} 
\cmidrule(lr){4-5} 
\cmidrule(lr){6-7} 
\cmidrule(lr){8-9}
& \cmarkgreen & \xmarkred 
& \cmarkgreen & \xmarkred  
& \cmarkgreen & \xmarkred  
& \cmarkgreen & \xmarkred  \\
\midrule
SkyGP-Fast-MoE ($\bar{\mathcal{N}}=1$) & \textbf{0.26} & 0.64 & - & - & \textbf{0.26} & \textbf{0.26} & \textbf{0.26} & \textbf{0.26} \\
SkyGP-Fast-MoE ($\bar{\mathcal{N}}=2$) & \textbf{0.18} & 0.35 & - & - & \textbf{0.18} & \textbf{0.18} & \textbf{0.18} & \textbf{0.18}\\
SkyGP-Fast-MoE ($\bar{\mathcal{N}}=4$) & \textbf{0.16} & 0.20 & - & - & \textbf{0.16} & \textbf{0.16} & \textbf{0.16} & \textbf{0.16}\\
SkyGP-Dense-MoE ($\bar{\mathcal{N}}=1$) & \textbf{0.23} & 0.55 & \textbf{0.23} & \textbf{0.23} & \textbf{0.23} & \textbf{0.23} & \textbf{0.23} & \textbf{0.23} \\
SkyGP-Dense-MoE ($\bar{\mathcal{N}}=2$) & \textbf{0.15} & 0.28 & \textbf{0.15} & 0.16 & \textbf{0.15} & \textbf{0.15} & \textbf{0.15}&  \textbf{0.15}\\
SkyGP-Dense-MoE ($\bar{\mathcal{N}}=4$) & \textbf{0.14} & 0.18 & \textbf{0.14} & \textbf{0.14} & \textbf{0.14} & \textbf{0.14} &\textbf{0.14}& \textbf{0.14}\\
SkyGP-Fast-gPoE($\bar{\mathcal{N}}=1$) & \textbf{0.26} & 0.64 & - & - & \textbf{0.26} & \textbf{0.26} & \textbf{0.26} & \textbf{0.26} \\
SkyGP-Fast-gPoE ($\bar{\mathcal{N}}=2$) & \textbf{0.13} & 0.22 & - & - & \textbf{0.13} & 0.14 & \textbf{0.13} & \textbf{0.13}\\
SkyGP-Fast-gPoE ($\bar{\mathcal{N}}=4$) & \textbf{0.09} & 0.10 & - & - & \textbf{0.09} & \textbf{0.09} & \textbf{0.09} & \textbf{0.09}\\
SkyGP-Dense-gPoE ($\bar{\mathcal{N}}=1$) & \textbf{0.22} & 0.55 & 0.23 & \textbf{0.22} & \textbf{0.23} & \textbf{0.23} &\textbf{0.22} & 0.023\\
SkyGP-Dense-gPoE ($\bar{\mathcal{N}}=2$) & \textbf{0.11} & 0.17 & \textbf{0.11} & 0.12 & \textbf{0.11} & \textbf{0.11} &\textbf{0.11} & \textbf{0.11} \\
SkyGP-Dense-PoE ($\bar{\mathcal{N}}=4$) & \textbf{0.08} & 0.10 & \textbf{0.08} & 0.09 & 0.08 & \textbf{0.08} &\textbf{0.08}& \textbf{0.09}\\
SkyGP-Fast-rBCM ($\bar{\mathcal{N}}=1$) & \textbf{0.26} & 0.64 & - & - & \textbf{0.26} & \textbf{0.26} & \textbf{0.26} & \textbf{0.26} \\
SkyGP-Fast-rBCM ($\bar{\mathcal{N}}=2$) & \textbf{0.12} & 0.20 & - & - & \textbf{0.12} & \textbf{0.12} & \textbf{0.12} & \textbf{0.12}\\
SkyGP-Fast-rBCM ($\bar{\mathcal{N}}=4$) & \textbf{0.09} & 0.11 & - & - & \textbf{0.09} & \textbf{0.09} & \textbf{0.09} & \textbf{0.09}\\
SkyGP-Dense-rBCM($\bar{\mathcal{N}}=1$) & \textbf{0.22} & 0.55 & 0.23 & \textbf{0.22} & \textbf{0.23} & \textbf{0.23} &\textbf{0.23} & \textbf{0.23} \\
SkyGP-Dense-rBCM ($\bar{\mathcal{N}}=2$) & \textbf{0.10} & 0.16 & \textbf{0.11} & \textbf{0.11} & 0.11 & \textbf{0.10} &\textbf{0.10}& 0.11\\
SkyGP-Dense-rBCM ($\bar{\mathcal{N}}=4$) & \textbf{0.08} & 0.10 & \textbf{0.09} & \textbf{0.09} & \textbf{0.09} & \textbf{0.09} &\textbf{0.08}& 0.09\\
\bottomrule
\end{tabular}
\end{table*}

\begin{table*}[t]
\centering
\caption{Average MSLL on the PUMA dataset. \cmarkgreen indicates that the strategy is used; \xmarkred indicates that it is not.}
\begin{tabular}{lcccccccc}
\toprule
{Strategy} 
& \multicolumn{2}{c}{{KD}} 
& \multicolumn{2}{c}{{DC}} 
& \multicolumn{2}{c}{{AW}} 
& \multicolumn{2}{c}{{TD}} \\
\cmidrule(lr){2-3} 
\cmidrule(lr){4-5} 
\cmidrule(lr){6-7} 
\cmidrule(lr){8-9}
& \cmarkgreen & \xmarkred 
& \cmarkgreen & \xmarkred  
& \cmarkgreen & \xmarkred  
& \cmarkgreen & \xmarkred  \\
\midrule
SkyGP-Fast-MoE ($\bar{\mathcal{N}}=1$) & \textbf{-0.92} & -0.64 & - & - & \textbf{-0.92} & \textbf{-0.92} & \textbf{-0.92} & \textbf{-0.92} \\
SkyGP-Fast-MoE ($\bar{\mathcal{N}}=2$) & \textbf{-1.00} & -0.70 & - & - & \textbf{-1.00} & \textbf{-1.00} & \textbf{-1.00} & \textbf{-1.00} \\
SkyGP-Fast-MoE ($\bar{\mathcal{N}}=4$) & \textbf{-1.03} & -0.82 & - & - & \textbf{-1.03} & \textbf{-1.03} & \textbf{-1.03} & \textbf{-1.03} \\
SkyGP-Dense-MoE ($\bar{\mathcal{N}}=1$) & \textbf{-1.02} & -0.69 & \textbf{-1.02} & \textbf{-1.02} & \textbf{-1.02} & \textbf{-1.02} & \textbf{-1.02}& \textbf{-1.02} \\
SkyGP-Dense-MoE ($\bar{\mathcal{N}}=2$) & \textbf{-1.10} & -0.87 & \textbf{-1.10} & \textbf{-1.10} & \textbf{-1.10} & \textbf{-1.10} &\textbf{-1.10} & \textbf{-1.10} \\
SkyGP-Dense-MoE ($\bar{\mathcal{N}}=4$) & \textbf{-1.09} & -0.93 & \textbf{-1.09} & -1.07 & \textbf{-1.09} & \textbf{-1.09} &\textbf{-1.09}& \textbf{-1.09} \\
SkyGP-Fast-gPoE($\bar{\mathcal{N}}=1$) & \textbf{-0.92} & -0.64 & - & - & \textbf{-0.92} & \textbf{-0.92} & \textbf{-0.92} & \textbf{-0.92} \\
SkyGP-Fast-gPoE ($\bar{\mathcal{N}}=2$) & \textbf{-1.15} & -0.98 & - & - & \textbf{-1.15} & \textbf{-1.15} & \textbf{-1.15} & \textbf{-1.15} \\
SkyGP-Fast-gPoE ($\bar{\mathcal{N}}=4$) & \textbf{-1.38} & -1.28 & - & - & \textbf{-1.37} & \textbf{-1.37} & \textbf{-1.38} & \textbf{-1.38}\\
SkyGP-Dense-gPoE ($\bar{\mathcal{N}}=1$) & \textbf{-1.02} & -0.69 & \textbf{-1.02} & \textbf{-1.02} & \textbf{-1.02} & -1.01 &\textbf{-1.02} & \textbf{-1.02} \\
SkyGP-Dense-gPoE ($\bar{\mathcal{N}}=2$) & \textbf{-1.22} & -1.04 & \textbf{-1.22} & \textbf{-1.22} & \textbf{-1.22} & \textbf{-1.22} &\textbf{-1.22}& \textbf{-1.22}\\
SkyGP-Dense-gPoE ($\bar{\mathcal{N}}=4$) & \textbf{-1.39} & -1.31 & \textbf{-1.39} & -1.38 &\textbf{-1.39} & \textbf{-1.39} &\textbf{-1.39}& \textbf{-1.39}\\
SkyGP-Fast-rBCM ($\bar{\mathcal{N}}=1$) & \textbf{-0.92} & -0.64 & - & - & \textbf{-0.92} & \textbf{-0.92} & \textbf{-0.92} & \textbf{-0.92} \\
SkyGP-Fast-rBCM ($\bar{\mathcal{N}}=2$) & \textbf{-1.26} & -1.09 & - & - & \textbf{-1.26} & \textbf{-1.26} & \textbf{-1.26} & \textbf{-1.26}\\
SkyGP-Fast-rBCM ($\bar{\mathcal{N}}=4$) & \textbf{-1.39} & -1.26 & - & - & \textbf{-1.39} & \textbf{-1.39} & \textbf{-1.39} & \textbf{-1.39}\\
SkyGP-Dense-rBCM($\bar{\mathcal{N}}=1$) & \textbf{-1.01} & -0.69 & \textbf{-1.02} & \textbf{-1.02} & \textbf{-1.01} & \textbf{-1.01} & \textbf{-1.01} & \textbf{-1.01} \\
SkyGP-Dense-rBCM ($\bar{\mathcal{N}}=2$) & \textbf{-1.31} & -1.16 & -1.31 & \textbf{1.32} & \textbf{1.32} & \textbf{-1.32} &\textbf{-1.31} & \textbf{-1.31} \\
SkyGP-Dense-rBCM ($\bar{\mathcal{N}}=4$) & \textbf{-1.40} & -1.30 & \textbf{-1.40} & -1.39 & \textbf{-1.40} & \textbf{-1.40}&\textbf{-1.40}& \textbf{-1.40}\\
\bottomrule
\end{tabular}
\end{table*}

\begin{table*}[t]
\centering
\caption{Average SMSE on the KIN40K dataset. \cmarkgreen indicates that the strategy is used; \xmarkred indicates that it is not.}
\begin{tabular}{lcccccccc}
\toprule
{Strategy} 
& \multicolumn{2}{c}{{KD}} 
& \multicolumn{2}{c}{{DC}} 
& \multicolumn{2}{c}{{AW}} 
& \multicolumn{2}{c}{{TD}} \\
\cmidrule(lr){2-3} 
\cmidrule(lr){4-5} 
\cmidrule(lr){6-7} 
\cmidrule(lr){8-9}
& \cmarkgreen & \xmarkred 
& \cmarkgreen & \xmarkred  
& \cmarkgreen & \xmarkred  
& \cmarkgreen & \xmarkred  \\
\midrule
SkyGP-Fast-MoE ($\bar{\mathcal{N}}=1$) & \textbf{0.62} & 0.66 & - & - & 0.67 & \textbf{0.62} & \textbf{0.61} & 0.62 \\
SkyGP-Fast-MoE ($\bar{\mathcal{N}}=2$) & \textbf{0.46} & 0.51 & - & - & 0.52 & \textbf{0.46} & \textbf{0.46} & \textbf{0.46}\\
SkyGP-Fast-MoE ($\bar{\mathcal{N}}=4$) & \textbf{0.46} & 0.47 & - & - & 0.51 & \textbf{0.46} & \textbf{0.46} & \textbf{0.46}\\
SkyGP-Dense-MoE ($\bar{\mathcal{N}}=1$) & \textbf{0.54} & 0.55 & \textbf{0.54} & 0.56 & 0.59 & \textbf{0.54} & \textbf{0.54}& \textbf{0.54}\\
SkyGP-Dense-MoE ($\bar{\mathcal{N}}=2$) & \textbf{0.44} & \textbf{0.44} & \textbf{0.44} & \textbf{0.44} & 0.49 & \textbf{0.44}& \textbf{0.44}& \textbf{0.44}\\
SkyGP-Dense-MoE ($\bar{\mathcal{N}}=4$) & \textbf{0.48} & 0.50 & 0.48 & \textbf{0.47} & 0.51 & \textbf{0.48} & \textbf{0.48}& \textbf{0.48}\\
SkyGP-Fast-gPoE($\bar{\mathcal{N}}=1$) & \textbf{0.62} & 0.66 & - & - & 0.67 & \textbf{0.62}  & \textbf{0.62} & \textbf{0.62} \\
SkyGP-Fast-gPoE ($\bar{\mathcal{N}}=2$) & \textbf{0.46} & 0.50 & - & - & 0.54 & \textbf{0.46} & \textbf{0.46} & \textbf{0.46}\\
SkyGP-Fast-gPoE ($\bar{\mathcal{N}}=4$) & \textbf{0.36} & 0.37 & - & - & 0.41 & \textbf{0.36} & \textbf{0.36} & \textbf{0.36}\\
SkyGP-Dense-gPoE ($\bar{\mathcal{N}}=1$) & \textbf{0.54} & 0.55 & \textbf{0.54} & 0.56 & 0.59 & \textbf{0.54} & \textbf{0.54}& \textbf{0.54}\\
SkyGP-Dense-gPoE ($\bar{\mathcal{N}}=2$) & 0.43 & \textbf{0.42} &  \textbf{0.43} & \textbf{0.43} & 0.48 & \textbf{0.43} & \textbf{0.43}& \textbf{0.43}\\
SkyGP-Dense-gPoE ($\bar{\mathcal{N}}=4$) & \textbf{0.38} & 0.39 & 0.38 & \textbf{0.37} & 0.41 & \textbf{0.38} &\textbf{0.38}& \textbf{0.38}\\
SkyGP-Fast-rBCM ($\bar{\mathcal{N}}=1$) & \textbf{0.62} & 0.66 & - & - & 0.67 & \textbf{0.62} & \textbf{0.62} & \textbf{0.62} \\
SkyGP-Fast-rBCM ($\bar{\mathcal{N}}=2$) & \textbf{0.42} & 0.46 & - & - & 0.49 & \textbf{0.42} & \textbf{0.42} & \textbf{0.42}\\
SkyGP-Fast-rBCM ($\bar{\mathcal{N}}=4$) & \textbf{0.35} & 0.37 & - & - & 0.40 & \textbf{0.35} & \textbf{0.35} & \textbf{0.35}\\
SkyGP-Dense-rBCM($\bar{\mathcal{N}}=1$) & \textbf{0.54} & 0.55 & \textbf{0.54} & 0.56 & 0.59 & \textbf{0.54}& \textbf{0.54}& \textbf{0.54}\\
SkyGP-Dense-rBCM ($\bar{\mathcal{N}}=2$) & \textbf{0.39} & \textbf{0.39} & \textbf{0.39} & \textbf{0.39} & 0.44 & \textbf{0.39} &\textbf{0.39}& \textbf{0.39}\\
SkyGP-Dense-rBCM ($\bar{\mathcal{N}}=4$) & \textbf{0.38} & 0.39 & 0.38 & \textbf{0.37} & 0.40 & \textbf{0.38}&\textbf{0.38}& \textbf{0.38}\\
\bottomrule
\end{tabular}
\end{table*}

\begin{table*}[t]
\centering
\caption{Average MSLL on the KIN40K dataset. \cmarkgreen indicates that the strategy is used; \xmarkred indicates that it is not.}
\begin{tabular}{lcccccccc}
\toprule
{Strategy} 
& \multicolumn{2}{c}{{KD}} 
& \multicolumn{2}{c}{{DC}} 
& \multicolumn{2}{c}{{AW}} 
& \multicolumn{2}{c}{{TD}} \\
\cmidrule(lr){2-3} 
\cmidrule(lr){4-5} 
\cmidrule(lr){6-7} 
\cmidrule(lr){8-9}
& \cmarkgreen & \xmarkred 
& \cmarkgreen & \xmarkred  
& \cmarkgreen & \xmarkred  
& \cmarkgreen & \xmarkred  \\
\midrule
SkyGP-Fast-MoE ($\bar{\mathcal{N}}=1$) & \textbf{-0.31} & -0.28 & - & - & -0.26 & \textbf{-0.31} & \textbf{-0.31}& \textbf{-0.31}\\
SkyGP-Fast-MoE ($\bar{\mathcal{N}}=2$) & \textbf{-0.43} & -0.39 & - & - & -0.36 & \textbf{-0.43} & \textbf{-0.43} & \textbf{-0.43} \\
SkyGP-Fast-MoE ($\bar{\mathcal{N}}=4$) & \textbf{-0.42} & -0.40 & - & - & -0.36 & \textbf{-0.42} & \textbf{-0.42}  & \textbf{-0.42}\\
SkyGP-Dense-MoE ($\bar{\mathcal{N}}=1$) & \textbf{-0.40} & \textbf{-0.40} & \textbf{-0.40} & -0.38 & -0.34 & \textbf{-0.40} & \textbf{-0.40}& \textbf{-0.40}\\
SkyGP-Dense-MoE ($\bar{\mathcal{N}}=2$) & \textbf{-0.47} & -0.46 & \textbf{-0.47} & \textbf{-0.47} & -0.40 & \textbf{-0.47} &\textbf{-0.47}& \textbf{-0.47}\\
SkyGP-Dense-MoE ($\bar{\mathcal{N}}=4$) & \textbf{-0.39} & -0.36 & -0.39 & \textbf{-0.42} & -0.35 & \textbf{-0.39} &\textbf{-0.39}& \textbf{-0.39}\\
SkyGP-Fast-gPoE($\bar{\mathcal{N}}=1$) & \textbf{-0.31} & -0.28 & - & - & -0.26 & \textbf{-0.31}  & \textbf{-0.31}  & \textbf{-0.31}\\
SkyGP-Fast-gPoE ($\bar{\mathcal{N}}=2$) & \textbf{-0.20} & -0.17 & - & - & -0.09 & \textbf{-0.20} & \textbf{-0.20}& \textbf{-0.20}\\
SkyGP-Fast-gPoE ($\bar{\mathcal{N}}=4$) & \textbf{-0.57} & -0.55 & - & - & -0.49 & \textbf{-0.57} & \textbf{-0.58} & \textbf{-0.58}\\
SkyGP-Dense-gPoE ($\bar{\mathcal{N}}=1$) & \textbf{-0.40} & \textbf{-0.40} & \textbf{-0.40} & -0.38 & -0.34 & \textbf{-0.40} & \textbf{-0.40}& \textbf{-0.40}\\
SkyGP-Dense-gPoE ($\bar{\mathcal{N}}=2$) & \textbf{-0.28} & -0.25 & -0.28 & \textbf{-0.29} & -0.20 & \textbf{-0.28} & \textbf{-0.28}& \textbf{-0.28}\\
SkyGP-Dense-gPoE ($\bar{\mathcal{N}}=4$) & \textbf{-0.57} & -0.55 & -0.57 & \textbf{-0.61} & -0.52 & \textbf{-0.57} & \textbf{-0.57}& \textbf{-0.57}\\
SkyGP-Fast-rBCM ($\bar{\mathcal{N}}=1$) & \textbf{-0.31} & -0.28 & - & - & -0.26 & \textbf{-0.31} & \textbf{-0.31} & \textbf{-0.31} \\
SkyGP-Fast-rBCM ($\bar{\mathcal{N}}=2$) & \textbf{-0.52} & -0.49 & - & - & -0.44 & \textbf{-0.52} & \textbf{-0.53} & \textbf{-0.53}\\
SkyGP-Fast-rBCM ($\bar{\mathcal{N}}=4$) & \textbf{-0.59} & -0.58 & - & - & -0.35 & \textbf{-0.50} & \textbf{-0.59} & \textbf{-0.59}\\
SkyGP-Dense-rBCM($\bar{\mathcal{N}}=1$) & \textbf{-0.40} & \textbf{-0.40} & \textbf{-0.40} & -0.38 & -0.34 & \textbf{-0.40} &\textbf{-0.40}& \textbf{-0.40}\\
SkyGP-Dense-rBCM ($\bar{\mathcal{N}}=2$) & \textbf{-0.59} & \textbf{-0.59} & -0.59 & \textbf{-0.60} & -0.51 & \textbf{-0.59}&\textbf{-0.59}& \textbf{-0.59}\\
SkyGP-Dense-rBCM ($\bar{\mathcal{N}}=4$) & \textbf{-0.61} & -0.60 & -0.61 & \textbf{-0.62} & -0.56 & \textbf{-0.61}&\textbf{-0.61}& \textbf{-0.61}\\
\bottomrule
\end{tabular}
\end{table*}

\begin{table*}[t]
\centering
\caption{Average SMSE on the ELECTRIC dataset. \cmarkgreen indicates that the strategy is used; \xmarkred indicates that it is not.}
\begin{tabular}{lcccccccc}
\toprule
{Strategy} 
& \multicolumn{2}{c}{{KD}} 
& \multicolumn{2}{c}{{DC}} 
& \multicolumn{2}{c}{{AW}} 
& \multicolumn{2}{c}{{TD}} \\
\cmidrule(lr){2-3} 
\cmidrule(lr){4-5} 
\cmidrule(lr){6-7} 
\cmidrule(lr){8-9}
& \cmarkgreen & \xmarkred 
& \cmarkgreen & \xmarkred  
& \cmarkgreen & \xmarkred  
& \cmarkgreen & \xmarkred  \\
\midrule
    SkyGP-Fast-MoE ($\bar{\mathcal{N}}=1$) & \textbf{0.08} & 0.10 & - & - & \textbf{0.08} & \textbf{0.08} & \textbf{0.08} & \textbf{0.08} \\
SkyGP-Fast-MoE ($\bar{\mathcal{N}}=2$) & \textbf{0.13} & 0.15 & - & - & \textbf{0.13} & \textbf{0.13} & \textbf{0.13} & \textbf{0.13} \\
SkyGP-Fast-MoE ($\bar{\mathcal{N}}=4$) & \textbf{0.17} & 0.22 & - & - & \textbf{0.17} & \textbf{0.17} & \textbf{0.17} & \textbf{0.17} \\
SkyGP-Dense-MoE ($\bar{\mathcal{N}}=1$) & \textbf{0.08} & 0.10 & \textbf{0.08} & \textbf{0.08} & \textbf{0.08} & \textbf{0.08} & \textbf{0.08}& \textbf{0.08}\\
SkyGP-Dense-MoE ($\bar{\mathcal{N}}=2$) & \textbf{0.12} & 0.15 & \textbf{0.12} & 0.13 & \textbf{0.12} & \textbf{0.12} &\textbf{0.12}& \textbf{0.12}\\
SkyGP-Dense-MoE ($\bar{\mathcal{N}}=4$) & \textbf{0.18} & 0.22 & \textbf{0.18} & \textbf{0.18} & \textbf{0.18} & \textbf{0.18} &\textbf{0.18} & \textbf{0.18}\\
SkyGP-Fast-gPoE($\bar{\mathcal{N}}=1$) & \textbf{0.08} & 0.11 & - & - & \textbf{0.08} & \textbf{0.08} & \textbf{0.08} & \textbf{0.08} \\
SkyGP-Fast-gPoE ($\bar{\mathcal{N}}=2$) & \textbf{0.09} & \textbf{0.09} & - & - & \textbf{0.09} & \textbf{0.09} & \textbf{0.09} & \textbf{0.09}\\
SkyGP-Fast-gPoE ($\bar{\mathcal{N}}=4$) & \textbf{0.08} & 0.09 & - & - & \textbf{0.08} & \textbf{0.08} & \textbf{0.08} & \textbf{0.08}\\
SkyGP-Dense-gPoE ($\bar{\mathcal{N}}=2$) & \textbf{0.07} & 0.09 & \textbf{0.07} & \textbf{0.07} & \textbf{0.07} & \textbf{0.07} & \textbf{0.07} & \textbf{0.07}\\
SkyGP-Dense-gPoE ($\bar{\mathcal{N}}=4$) & \textbf{0.07} & 0.08 & \textbf{0.07} & 0.08 & \textbf{0.07} & \textbf{0.07} &\textbf{0.07}& \textbf{0.07}\\
SkyGP-Fast-rBCM ($\bar{\mathcal{N}}=1$) & \textbf{0.08} & 0.11 & - & - & \textbf{0.08} & 0.09  & \textbf{0.08} & \textbf{0.08} \\
SkyGP-Fast-rBCM ($\bar{\mathcal{N}}=2$) & \textbf{0.09} & \textbf{0.09} & - & - & 0.09 & \textbf{0.08} & \textbf{0.09} & \textbf{0.09}\\
SkyGP-Fast-rBCM ($\bar{\mathcal{N}}=4$) & \textbf{0.07} & 0.09 & - & - & \textbf{0.07} & \textbf{0.07} & \textbf{0.07} & \textbf{0.07}\\
SkyGP-Dense-rBCM($\bar{\mathcal{N}}=1$) & \textbf{0.08} & 0.10 & \textbf{0.08} & \textbf{0.08} & \textbf{0.08} & 0.09 & \textbf{0.08}& \textbf{0.08}\\
SkyGP-Dense-rBCM ($\bar{\mathcal{N}}=2$) & \textbf{0.07} & 0.09 & \textbf{0.07} & \textbf{0.07} & \textbf{0.07} & 0.08 & \textbf{0.07}& \textbf{0.07}\\
SkyGP-Dense-rBCM ($\bar{\mathcal{N}}=4$) & \textbf{0.07} & 0.08 & \textbf{0.07} & 0.08 & \textbf{0.07} & \textbf{0.07} & \textbf{0.07}& \textbf{0.07}\\
\bottomrule
\end{tabular}
\end{table*}

\begin{table*}[t]
\centering
\caption{Average MSLL on the ELECTRIC dataset. \cmarkgreen indicates that the strategy is used; \xmarkred indicates that it is not.}
\begin{tabular}{lcccccccc}
\toprule
{Strategy} 
& \multicolumn{2}{c}{{KD}} 
& \multicolumn{2}{c}{{DC}} 
& \multicolumn{2}{c}{{AW}} 
& \multicolumn{2}{c}{{TD}} \\
\cmidrule(lr){2-3} 
\cmidrule(lr){4-5} 
\cmidrule(lr){6-7} 
\cmidrule(lr){8-9}
& \cmarkgreen & \xmarkred 
& \cmarkgreen & \xmarkred  
& \cmarkgreen & \xmarkred  
& \cmarkgreen & \xmarkred  \\
\midrule
SkyGP-Fast-MoE ($\bar{\mathcal{N}}=1$) & \textbf{-3.30} & -3.12 & - & - & -3.29 & \textbf{-3.30} & \textbf{-3.30} & \textbf{-3.30} \\
SkyGP-Fast-MoE ($\bar{\mathcal{N}}=2$) & \textbf{-2.73} & -2.45 & - & - & \textbf{-2.73} & \textbf{-2.73} & \textbf{-2.73} & \textbf{-2.73}\\
SkyGP-Fast-MoE ($\bar{\mathcal{N}}=4$) & \textbf{-2.40} & -2.09 & - & - & \textbf{-2.40} & \textbf{-2.40} & \textbf{-2.40} & \textbf{-2.40}\\
SkyGP-Dense-MoE ($\bar{\mathcal{N}}=1$) & \textbf{-3.32} & -3.18 & \textbf{-3.32} & -3.31 & \textbf{-3.32} & \textbf{-3.32} & \textbf{-3.32}& \textbf{-3.32}\\
SkyGP-Dense-MoE ($\bar{\mathcal{N}}=2$) & \textbf{-2.78} & -2.50 & -2.78 & \textbf{-2.79} & -2.78 & \textbf{-2.79} &\textbf{-2.78}& \textbf{-2.78}\\
SkyGP-Dense-MoE ($\bar{\mathcal{N}}=4$) & \textbf{-2.38} & -2.06 & -2.38 & \textbf{-2.41} & \textbf{-2.38} & \textbf{-2.38} &\textbf{-2.38}& \textbf{-2.38}\\
SkyGP-Fast-gPoE($\bar{\mathcal{N}}=1$) & \textbf{-3.30} & -3.12 & - & - & \textbf{-3.30} & -3.28 & \textbf{-3.30} & \textbf{-3.30} \\
SkyGP-Fast-gPoE ($\bar{\mathcal{N}}=2$) & \textbf{-3.16} & -2.99 & - & - & \textbf{-3.17} & \textbf{-3.17} & \textbf{-3.16} & \textbf{-3.16}\\
SkyGP-Fast-gPoE ($\bar{\mathcal{N}}=4$) & \textbf{-3.29} & -3.18 & - & - & -3.29 & \textbf{-3.32} & \textbf{-3.29} & \textbf{-3.29}\\
SkyGP-Dense-gPoE ($\bar{\mathcal{N}}=1$) & \textbf{-3.32} & -3.18 & \textbf{-3.32} & -3.31 & \textbf{-3.31} & \textbf{-3.31} & \textbf{-3.32}& \textbf{-3.32}\\
SkyGP-Dense-gPoE ($\bar{\mathcal{N}}=2$) & \textbf{-3.24} & -3.14 & \textbf{-3.24} & -3.20 & \textbf{-3.23} & \textbf{-3.23} & \textbf{-3.24}& \textbf{-3.24}\\
SkyGP-Dense-gPoE ($\bar{\mathcal{N}}=4$) & \textbf{-3.30} & -3.26 & {-3.30} & \textbf{-3.31} & -3.30 & \textbf{-3.33} &\textbf{-3.30}& \textbf{-3.30}\\
SkyGP-Fast-rBCM ($\bar{\mathcal{N}}=1$) & \textbf{-3.30} & -3.11 & - & - & \textbf{-3.30} & -3.28 & \textbf{-3.30} & \textbf{-3.30} \\
SkyGP-Fast-rBCM ($\bar{\mathcal{N}}=2$) & \textbf{-3.30} & -3.20 & - & - & -3.30 & \textbf{-3.34} & \textbf{-3.30} & \textbf{-3.30}\\
SkyGP-Fast-rBCM ($\bar{\mathcal{N}}=4$) & \textbf{-3.35} & -3.28 & - & - & -3.35 & \textbf{-3.39} & \textbf{-3.35} & \textbf{-3.35}\\
SkyGP-Dense-rBCM($\bar{\mathcal{N}}=1$) & \textbf{-3.32} & -3.18 & \textbf{-3.32} & \textbf{-3.32} & \textbf{-3.32} & -3.31 &\textbf{-3.32}& \textbf{-3.32}\\
SkyGP-Dense-rBCM ($\bar{\mathcal{N}}=2$) & \textbf{-3.36} & -3.30 & \textbf{-3.36} & \textbf{-3.36} & \textbf{-3.36} & \textbf{-3.36} &\textbf{-3.36}& \textbf{-3.36}\\
SkyGP-Dense-rBCM ($\bar{\mathcal{N}}=4$) & \textbf{-3.36} & -3.34 & {-3.36} & \textbf{-3.37} & -3.36 & \textbf{-3.39}&\textbf{-3.36}& \textbf{-3.36}\\
\bottomrule
\end{tabular}
\end{table*}

\end{document}